\pdfoutput=1
\documentclass[11pt]{article}

\usepackage[preprint]{acl}

\usepackage{times}
\usepackage{latexsym}

\usepackage[T1]{fontenc}
\usepackage[utf8]{inputenc}

\usepackage{microtype}

\usepackage{inconsolata}

\usepackage{graphicx}
\usepackage{booktabs}
\usepackage{xcolor}
\usepackage{tcolorbox}
\usepackage{bbding}
\usepackage{pifont}
\usepackage[nointegrals]{wasysym}
\usepackage{amsmath}
\usepackage{amssymb}
\usepackage{footnote}

\newcommand{\Hquad}{\hspace{0.3em}} 
\newcommand{\ourbenchmark}{FRoG\Hquad}
\newcommand{\ourbenchmarknospace}{\textsc{FRoG}}
\newcommand{\benchmarkeasy}{FRoG-Easy}
\newcommand{\benchmarkhard}{FRoG-Hard}
\newcommand{\Thead}[1]{\textbf{\textsc{#1}}}

\newcommand{\quantfuzzyreason}[1]{\textcolor{teal}{#1}}

\newenvironment{itemize*}%
 {\leftmargini=10pt\begin{itemize}%
  \setlength{\itemsep}{0pt}%
  \setlength{\parskip}{0pt}%
  }%
 {\end{itemize}}
\newenvironment{enumerate*}%
 {\begin{enumerate}%
  \setlength{\itemsep}{0pt}%
  \setlength{\parskip}{0pt}}%
 {\end{enumerate}}

\title{\includegraphics[width=0.04\textwidth]{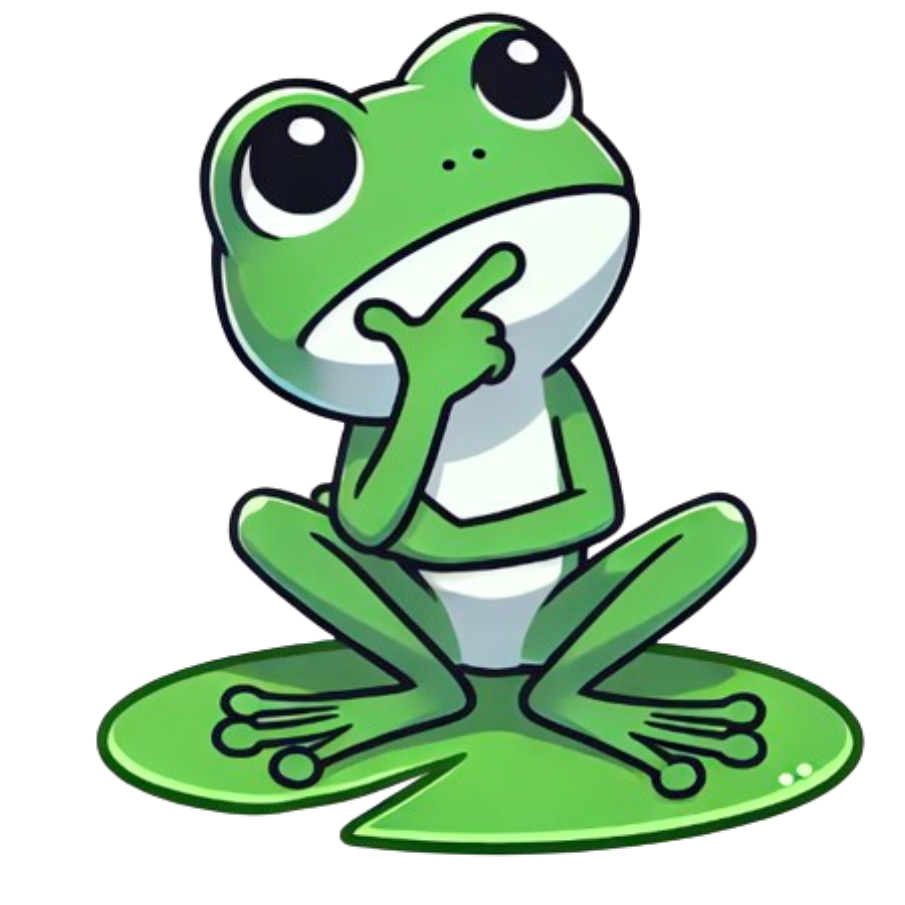}~\ourbenchmarknospace: Evaluating Fuzzy Reasoning of Generalized Quantifiers in Large Language Models}

\author{Yiyuan Li\textsuperscript{\includegraphics[scale=0.05]{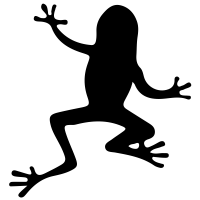}}  \qquad  Shichao Sun\textsuperscript{\includegraphics[scale=0.05]{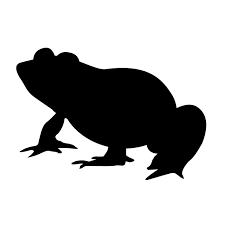}}  \qquad Pengfei Liu\textsuperscript{\includegraphics[scale=0.05]{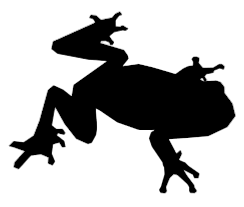}} \thanks{Corresponding author} \\
  Shanghai Jiao Tong University\textsuperscript{\includegraphics[scale=0.05]{figures/frogs/frog3.png}} \quad UNC-Chapel Hill\textsuperscript{\includegraphics[scale=0.05]{figures/frogs/frog1.png}}\\ The Hong Kong Polytechnic University\textsuperscript{\includegraphics[scale=0.05]{figures/frogs/frog2.png}}\\
  \texttt{yiyuanli@cs.unc.edu, csssun@comp.polyu.edu.hk, pengfei@sjtu.edu.cn}
}

\begin{document}
\maketitle
\begin{abstract}
Fuzzy reasoning is vital due to the frequent use of imprecise information in daily contexts. However, the ability of current large language models (LLMs) to handle such reasoning remains largely uncharted. In this paper, we introduce a new benchmark, \ourbenchmarknospace, for fuzzy reasoning, featuring real-world mathematical word problems that incorporate generalized quantifiers. Our experimental findings reveal that fuzzy reasoning continues to pose significant challenges for LLMs. Moreover, we find that existing methods designed to enhance reasoning do not consistently improve performance in tasks involving fuzzy logic. Additionally, our results show an inverse scaling effect in the performance of LLMs on \ourbenchmarknospace. Interestingly, we also demonstrate that strong mathematical reasoning skills are not necessarily indicative of success on our benchmark.\footnote{Resource: \url{https://github.com/Nativeatom/FRoG}}.
\end{abstract}

\section{Introduction}
The capability to reason effectively is a critical dimension in evaluating the advancements of large language models (LLMs)~\citep{NEURIPS2020_GPT3,huang2022towards,bowen-etal-2024-comprehensive,wang2024reasoning,zhu2024dyval}. Commonly, the progress of advancing reasoning abilities is mainly evaluated on mathematical reasoning benchmarks \citep{xia2024evaluating,huang2024key} like GSM8K~\cite{cobbe2021gsm8k} and MATH~\cite{hendrycksmath2021}, which demands \textbf{precise} answers derived from clear, rule-based questions. However, much of human knowledge and daily decision-making processes are not based on precision but rather involve handling \textbf{fuzzy}, imprecise information~\cite{Novák2015}.

Fuzzy reasoning, which deals with uncertainties and perceptual data, differs significantly from the deterministic processes evaluated in existing standard benchmarks. It often relies on natural language constructs that do not capture information with precise granularity~\cite{Novák2015}. For instance, generalized quantifiers (GQs), such as ``few'' or ``most'', are frequently used in natural language to introduce a level of vagueness \cite{Mostowski1957}. An illustrative fuzzy problem employing GQ might be: ``\textit{There have been successive increases of 20\% and then \underline{most} in the price of gas from the previous month. By what percentage should a driver reduce gas consumption so that the expenditure does not change?}'' Here, the term ``\emph{most}'' introduces ambiguity concerning the extent of the price increase and necessitates an estimation of its semantics to solve the problem accurately.

However, such GQ-based fuzzy reasoning problems are under-explored. \citet{faghihi2020association} introduces probabilistic fuzzy logic~\cite{fuzzylogic} to enhance reasoning abilities. % in real-life problems. 
Existing works like \citet{saki2022fundamental} investigate employing fuzzy probabilistic theory in association-based problems, like the relation between fuzzy features and labels (e.g. \underline{\textit{a lot of}} smoking and cancer). The mapping functions to encode fuzzy features are either simulated~\cite{faghihi2020association} or collected from limited data with heuristics~\cite{8788644}. Those mapping functions fail to evaluate complex, real-world reasoning scenarios and realistically capture how LLMs utilize fuzzy reasoning in practical contexts.  

\begin{figure*}[t!]
	\centering
	\includegraphics[width=1\textwidth]{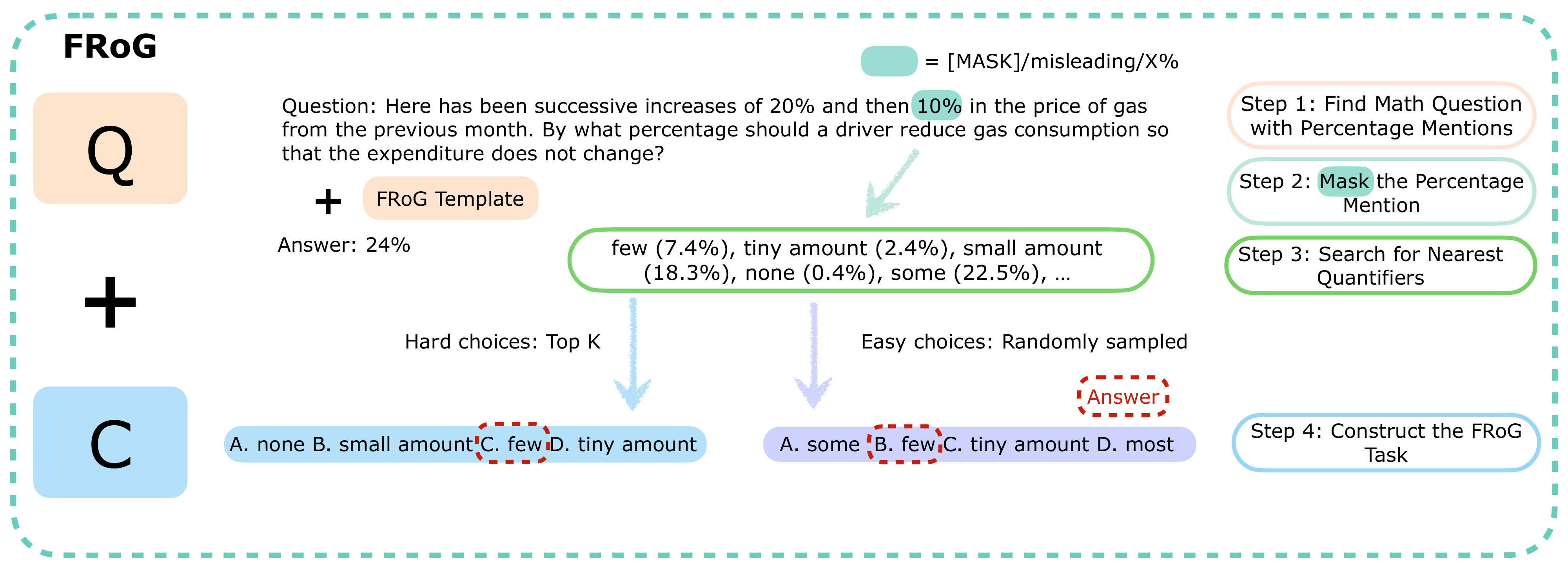}
	\caption{Workflow of \ourbenchmark construction.}
	\label{fig: frog_overview}
\end{figure*}

In this paper, we aim to explore reasoning challenges associated with fuzzy events~\cite{Zadeh1968ProbabilityMO}, which are mathematically ambiguous and articulated through fuzzy expressions of GQs, such as ``most in the price of gas''. Specifically, we focus on mathematical reasoning problems that incorporate GQs. To this end, we have developed \ourbenchmarknospace, a benchmark for \textbf{F}uzzy \textbf{R}easoning benchmark \textbf{o}f \textbf{G}eneralized quantifiers, which transforms real-world mathematical problems from GSM8K~\cite{cobbe2021gsm8k} and MathQA~\cite{amini-etal-2019-mathqa} datasets into multiple-choice questions. These questions replace precise numerical data with GQs, requiring models to apply both precise reasoning like arithmetic computations and fuzzy reasoning to estimate the scope of GQs. Our evaluation of several LLMs reveals that fuzzy reasoning remains a significant challenge in \ourbenchmarknospace, with an inverse scaling effect observed in over half of the model families tested. Furthermore, common methods designed to enhance reasoning capabilities like math-specialized tuning, code-specialized tuning, and general alignment, do not consistently improve performance in \ourbenchmarknospace. Interestingly, we find that strong mathematical reasoning skills are not necessarily indicative of success on \ourbenchmarknospace. We also outline the primary strategies employed by LLMs to perform fuzzy reasoning.

\section{Related Work}
Reasoning abilities, involving drawing conclusions from existing knowledge, are a cornerstone of human intelligence and are crucial for intricate tasks like decision-making and solving math word problems~\cite{yu2023natural}. Recently, mathematical problem-solving has become a key dimension in assessing the reasoning capabilities of LLMs~\cite{xia2024evaluating,huang2024key}. Various approaches have been developed to enhance the math reasoning ability of LLMs, like prompt-based designs~\cite{chia2023contrastive, zheng2023progressive, chen2023skills}. On the other hand, \citet{taylor2022galactica,lewkowycz2022solving,paster2023openwebmath,azerbayev2024llemma} propose data construction for the pertaining stage. And further supervised fine-tuning, instruction tuning or alignment methods like Direct Preference Optimization (DPO, ~\citet{rafailov2023direct}) are used to enhance the reasoning abilities of LLMs~\cite{yu2023metamath,luo2023wizardmath,an2023learning,huang2024mustard,li2024synthetic, deepseekai2024deepseek}. Motivated by the fact that external tools are widely used in NLP tasks, tool integration is introduced to enhance math reasoning~\cite{mishra2022lila,gao2023pal,gou2023critic,gou2023tora,yue2023mammoth,zhou2023solving,zhang2024evaluating}. \citet{chen2023program} employs programming methods to enhance the reasoning ability of LLMs. In \ourbenchmarknospace, we build the fuzzy reasoning tasks based on mathematical reasoning problems and explore common approaches designed to improve reasoning capabilities including math-specialized tuning, code-specialized tuning, and general alignment.

GQs are widely used to indicate proportions of predicate satisfaction in communication~\cite{joshi-etal-2020-taxinli} and NLP benchmarks~\cite{suhr-etal-2019-corpus, cui-etal-2022-generalized-quantifiers}. They also contribute as a major source to the deficiencies of NLP systems such as NLI~\cite{cui-etal-2022-generalized-quantifiers}. Given their prevalence in the real-world, we employ GQs as a natural approach to introduce fuzzy information in \ourbenchmarknospace.

Existing methods for modeling fuzzy logic in natural language, as developed by~\citet{lee2004first} and ~\citet{kapustin-kapustin-2019-modeling}, depend on pre-defined mapping functions to process fuzziness. Those mapping functions are mostly built from rule-based heuristics on limited data, or with simple distribution assumption (e.g. the gaussian distribution) and hard to be directly on real-world complex reasoning problems. In \ourbenchmarknospace, we rely on LLMs that are pretrained from large-scale real world text corpora to process the fuzziness of GQs and conduct complex math reasoning problems.

\section{Benchmark Collection}
Problems in \ourbenchmark are collected from two math word problem datasets from the real world: GSM8K and MathQA. GSM8K includes grade school math word problems that can be largely solved with basic arithmetic operations. MathQA consists of multiple-choice GRE and GMAT-level math problems. In \ourbenchmarknospace, we include questions with percentage mentions, and Figure~\ref{fig: frog_overview} displays an overview of the collection workflow. Specifically,

\begin{itemize*}
    \item \textbf{Step 1: Identifying Mathematical Questions with Percentage Mentions} - We begin by filtering the original questions to include only those that contain at least one percentage figure, of which the value is between $0\%$ and $100\%$.
    \item \textbf{Step 2: Masking the Percentage Mention} - We obscure the specific target percentage mention by replacing it with a [MASK] token to construct a \textit{Mask} question. If the original question contains multiple percentage mentions, each mention is masked out separately. We also employ two other masking strategies \textit{Mislead} and \textit{X\%} where the target mention is substituted with a misleading quantifier (with the incorrectness pointed out in the \ourbenchmark template) or X\%.
    \item \textbf{Step 3: Searching for the Nearest Quantifiers} - The golden choice is selected by finding the closest GQ according to its average strength provided in QuRe~\cite{li-etal-2023-pragmatic}, a quantifier reasoning dataset with human-annotated quantifier strengths (i.e., \textit{few} is the closest GQ to \textit{10\%} in the running example).
    \item \textbf{Step 4: Constructing the \ourbenchmark Task} - In \ourbenchmarknospace, we provide the question and the original answer to infer which GQ can be filled to represent the information masked out. This framework is driven by the acknowledgment that it is more practical to formulate fuzzy reasoning than directly solve mathematical problems with GQs.
    To carefully investigate the performance, we design the \textit{easy} and \textit{hard} mode of choices depending on the discriminability of misleading choices. The incorrect choices in \benchmarkhard \Hquad are the misleading top GQs in the previous step (e.g. \textit{tiny amount, small amount, none} in the running example), while incorrect choices in \benchmarkeasy \Hquad are randomly sampled from all misleading GQs. The original question, original answer and choices are then assembled through \ourbenchmark templates. We refer to Appendix~\ref{sec: FRoG templates} for details.
\end{itemize*}

For MathQA data, the original multiple choice format is omitted in \ourbenchmark since there is a negligible difference in performance between maintaining the original choice or providing the correct numeric answer, according to preliminary experiments. 

\begin{figure}[t!]
	\centering
        \includegraphics[width=0.5\textwidth]{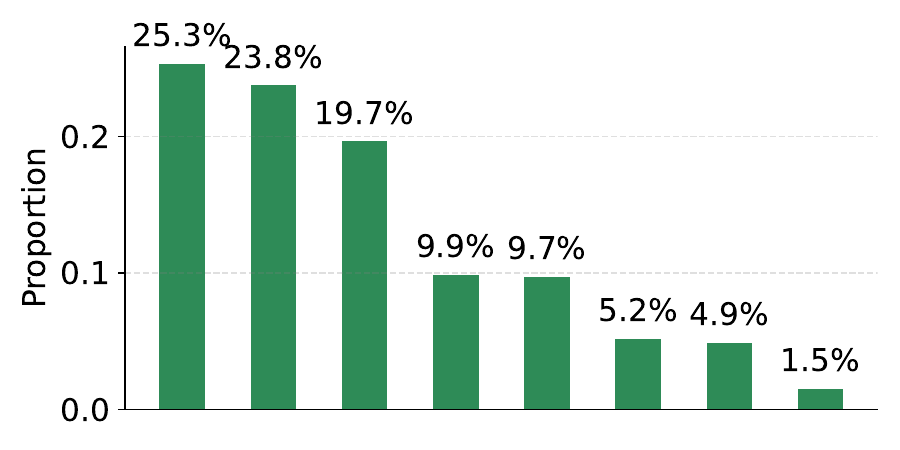}
        \includegraphics[width=0.5\textwidth]{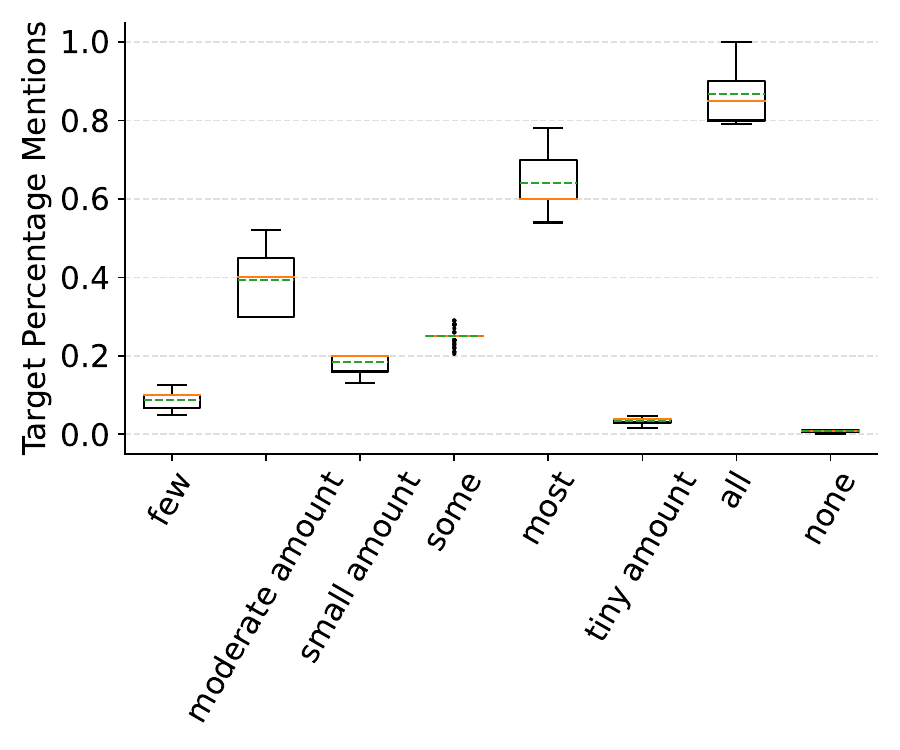}
	\caption{(Top) quantifier proportions in \ourbenchmarknospace. (Bottom) percentiles of target percentage mentions categorized by quantifiers. Green and orange lines represent the means and medians, respectively. The x-axis is shared between the two figures.}
	\label{fig: FRoG_quant_dist}
\end{figure}

Eventually, \ourbenchmark includes 199 questions from GSM8K and 1,845 questions from MathQA. The average number of question tokens is 68.2, with each token separated by a space. And each question in \ourbenchmark contains an average of 1.6 percentage mentions. The total number of quantifiers involved in \ourbenchmark is 8, and the most common quantifiers used are few (25.3\%), moderate amount (23.8\%), and small amount (19.7\%) (see top Figure~\ref{fig: FRoG_quant_dist} for details). The bottom of Figure~\ref{fig: FRoG_quant_dist} reveals the target percentage mentions mapped to each quantifier, e.g. the mean percentage value and 0.15 for \textit{few} and around 0.4 for \textit{moderate amount}.

\section{Experiment}

\begin{figure*}[t!]
	\centering
	\includegraphics[width=1\textwidth]{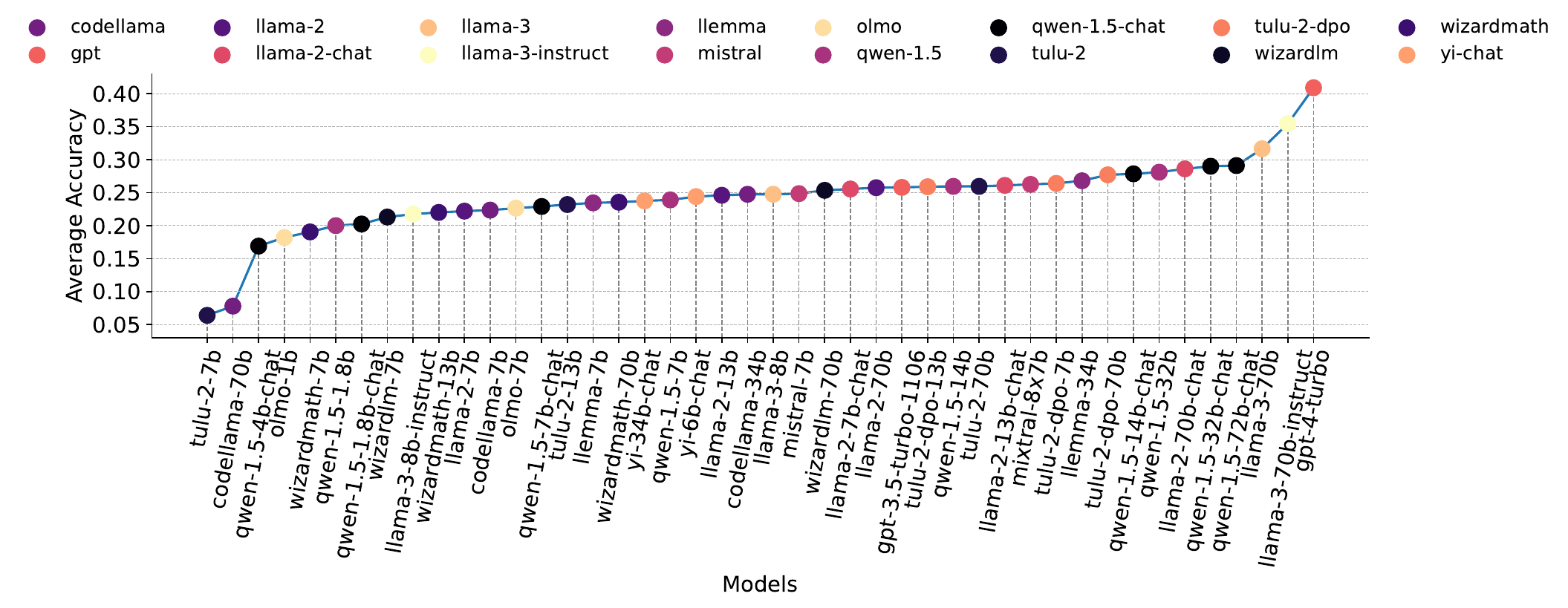}
	\caption{The average \textit{Mask} accuracy in \benchmarkeasy \Hquad and \benchmarkhard \Hquad of several LLMs sorting in ascending order. Dots with the same color belong to the same model family. Models with additional pretraining or instruction tuning do not necessarily perform better. We refer to Figure~\ref{fig: impact_math_train_mathqa} and Figure~\ref{fig: impact_alignment_mathqa} for more details.}
	\label{fig: results_frog}
\end{figure*}

We evaluated several open-sourced LLMs, including Llama-2~\cite{touvron2023llama}, CodeLlama~\cite{rozière2024code}, Llemma~\cite{azerbayev2024llemma}, llama-3~\cite{llama-3}, Mistral~\cite{jiang2023mistral}, Qwen-1.5~\cite{qwen}, Tulu-2~\cite{ivison2023camels}, WizardLM~\cite{xu2024wizardlm}, WizardMath~\cite{luo2023wizardmath} and Yi-Chat~\cite{young2024yi} on our \ourbenchmark benchmark.
Specifically, we would like to investigate
the following three research questions:
\begin{enumerate*}
\item Effectiveness of existing reasoning enhancement methods in \ourbenchmarknospace?\\

\item Can the scaling law be observed in \ourbenchmarknospace?\\

\item Is strong mathematical reasoning ability transferrable to fuzzy reasoning?
\end{enumerate*}

We employ the greedy decoding strategy with max tokens being 1,000, and temperature being 0.9 in experiments. The LLMs are instructed with task instructions (see Appendix~\ref{sec: FRoG instruction}) and 5 demonstrations~\cite{NEURIPS2020_GPT3} with manually created chain-of-thought~\cite{wei2022chain} solutions to assist the reasoning procedure. The experiments are conducted on NVIDIA A100 80GB GPUs, each experiment can be finished within 2 hours.

Moreover, we investigate the sensitivity of different masking strategies. Specifically, we compare the performance between the \textit{Mask} and \textit{Mislead} or \textit{X\%} task by computing the Pearson and Spearman correlation of their accuracy. The results in Table~\ref{tab: task correlation} indicate strong positive correlations between the performance of \textit{Mask} and \textit{Mislead} or \textit{X\%}, meaning that LLMs are not sensitive to the masking strategy in \ourbenchmarknospace. We also do not observe strong correlation between accuracy and length of the generation. We choose the \textit{Mask} task as the major task thereafter. 

\begin{table}[ht]
    \centering
    \begin{tabular}{c|c|c}
        \toprule
        \textbf{Task} & \textbf{Pearson} & 
        \textbf{Spearman} \\ 
        \midrule
        Mislead & 0.966/0.935 & 0.895/0.683 \\
        \midrule
        X\% & 0.988/0.980 & 0.958/0.889 \\
        \bottomrule
    \end{tabular}
    \caption{The Pearson and Spearman correlation between performance of different task and \textit{Mask}. The two values corresponds to \benchmarkeasy \Hquad and \benchmarkhard, and all p values are smaller than 0.01. The performance of \textit{Mislead} and \textit{X\%} are strongly correlated to \textit{Mask}.}
    \label{tab: task correlation}
\end{table}

\subsection{Overall Result}
The result on \ourbenchmark is displayed in Figure~\ref{fig: results_frog}. In general, the accuracy of all models is around 0.05 and 0.45 (and mostly between 0.15 and 0.3), indicating that the fuzzy reasoning is a challenging task for the current LLMs. Moreover, models with small model sizes can demonstrate suprisingly strong performance in \ourbenchmark compared to models much larger, e.g. Tulu-2-DPO-7B outperforms Tulu-2-70B, Llama-2-70B and WizardLM-70B.

\subsection{Fine-grained Evaluation and Analysis}

\subsubsection{Q1: Effectiveness of existing reasoning enhancement methods in \ourbenchmarknospace?}
Instruction-tuning~\cite{wei2022finetuned, instructgpt} is demonstrated to further boost the abilities of LLMs~\cite{zhang2023instruction,hu2024fine}. It is employed to improve the usability and safety of LLM systems (e.g., the chat model~\citet{touvron2023llama,wang2024survey}), open-ended generations without sacrificing task-specific abilities~\cite{ivison2023camels} or mathematical reasoning abilities~\cite{qwen,tang2024mathscale,zhou2024dual}. Moreover,
\citet{zhang2024unveiling} demonstrates consistent performance gain of LLMs on reasoning by instruction-tuning on code data, and the mathematical reasoning ability can be enhanced by introducing a continuous pretraining stage on mathematical tokens~\cite{luo2023wizardmath, azerbayev2024llemma}.

We explore whether the effectiveness of continuous pretraining on math or code data, as well as general alignment tuning methods can be extended to fuzzy reasoning in \ourbenchmarknospace. In this regard, we selectively compare the performance of sevel models WizardLM, Qwen-1.5, Tulu-2, Llama-2, and Llama-3, and their instruction-tuning or aligned versions: WizardMath (based on WizardLM), CodeLlama (based on Llama-2), and Llemma (based on CodeLlama).  

\begin{figure}[t!]
	\centering
        \includegraphics[width=0.5\textwidth]{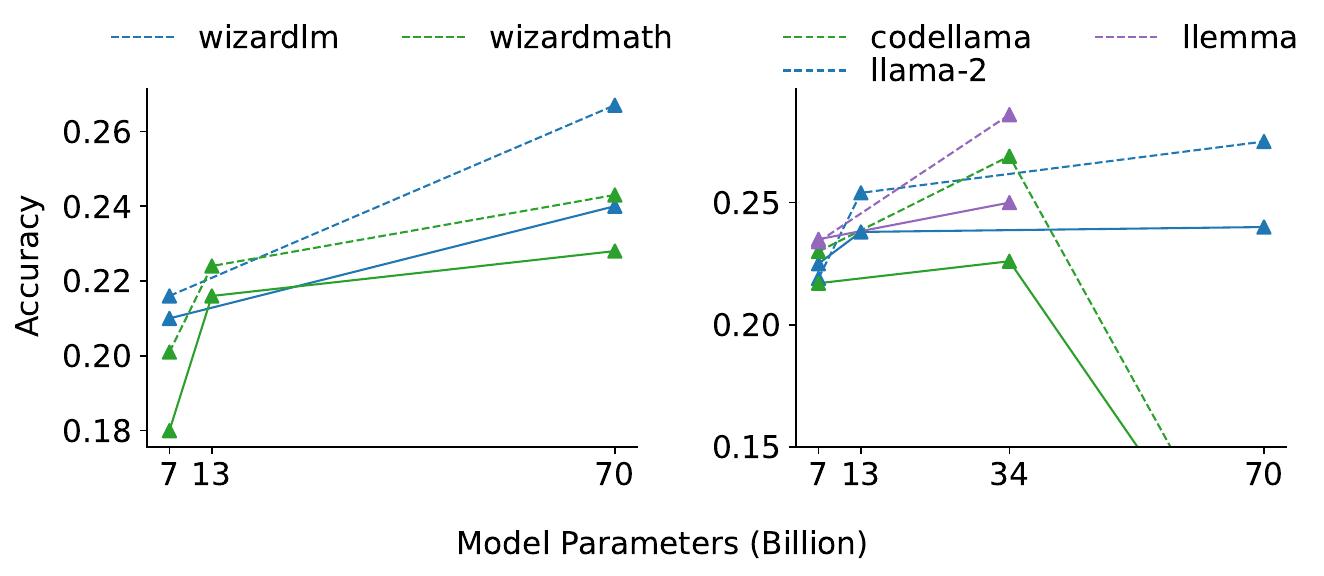}
	\caption{Impacts of continuous pretraining on mathematical data of LLMs on the performance of \ourbenchmarknospace. The solid and dashed lines represent \benchmarkhard \Hquad and \benchmarkeasy \Hquad respectively. The result of CodeLlama (70B) is emitted for illustration due to its poor performance.}
	\label{fig: impact_math_train_mathqa}
\end{figure}

\paragraph{Math-specialized Tuning} The results are demonstrated in Figure~\ref{fig: impact_math_train_mathqa}, where we observe that the accuracy of all LLMs are less than 30\% and the mathematical continuous training does not bring universal benefits in FRoG since WizardLM outperforms WizardMath in \ourbenchmarknospace. Besides, the scaling benefit of Llemma on the \benchmarkeasy \Hquad (5.2\% gain) does not comparably extend to the \benchmarkhard \Hquad (1.5\%).

\paragraph{Code-specialized Tuning} CodeLlama does not outperform Llama-2 in \ourbenchmarknospace, meaning continuous pretraining on code does not directly benefit fuzzy reasoning tasks. In fact, the domain shift from programming to natural language largely impacts the performance. The largest CodeLlama (70B) evaluated directly generates code snippet most of the time, leading to the poor performance on \ourbenchmarknospace.

\paragraph{General Alignment} 

\begin{figure*}[t!]
	\centering
        \includegraphics[width=\textwidth]{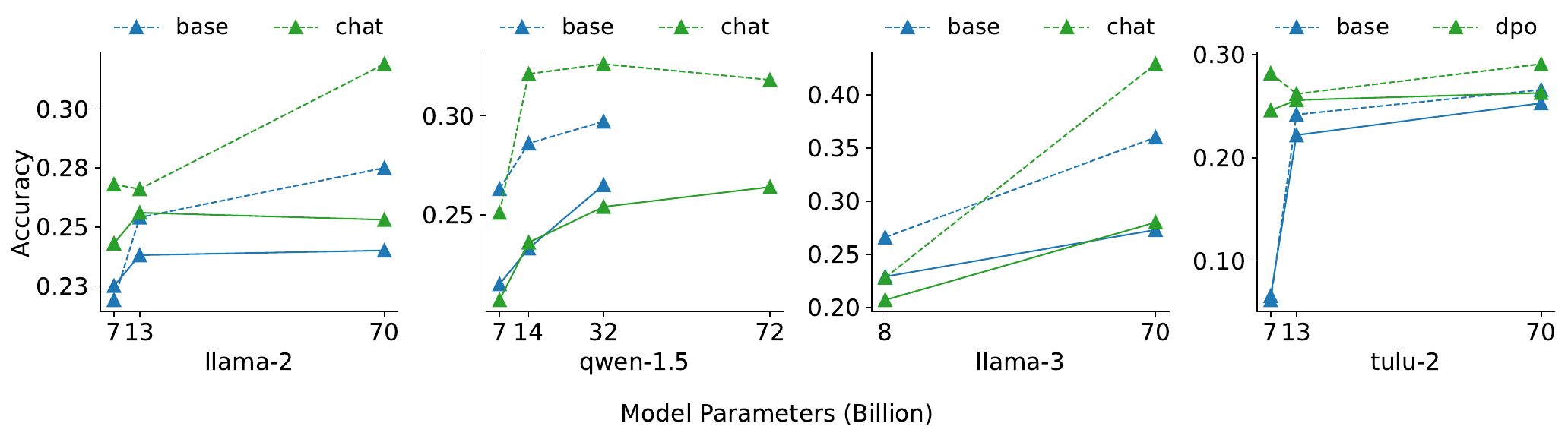}
	\caption{Comparison between different chat and base models of \textit{Mask} on \ourbenchmarknospace. The solid and dashed lines represent the \textit{hard} and \textit{random} modes, respectively. Instruction-tuning does not necessarily improve the performance in \ourbenchmark. The results of qwen-1.5-72b are full of punctuations and therefore omitted.}
	\label{fig: impact_alignment_mathqa}
\end{figure*}

The results are shown in Figure~\ref{fig: impact_alignment_mathqa}. 
In general, the performance on \benchmarkeasy \Hquad (dashed lines) is better than \benchmarkhard \Hquad (real lines) among all the models. Regarding the difference between the base models and their instruction-tuned versions, the benefit of instruction-tuning substantially diminishes from \benchmarkeasy \Hquad to \benchmarkhard \Hquad in Llama-2, Qwen-1.5 and Llama-3. Lastly, the inverse scaling effect displays on Llama-2, Qwen-1.5 and Tulu-2. In \benchmarkeasy, the perfromance of 4 base models adhere to the scaling law, whereas the instruction-tuned models display inverse scaling effect except Llama-3. In \benchmarkhard, Llama-2 demonstrates inverse scaling effect in both base and chat models. Conversely, Qwen-1.5 and Llama-3 consistently exhibit scaling phenomenon in both base and chat models. The Tulu-2 base model adheres to the scaling law, while the Tulu-2-DPO model display the inverse scaling effect.

\begin{figure}[t!]
	\centering
        \includegraphics[width=0.5\textwidth]{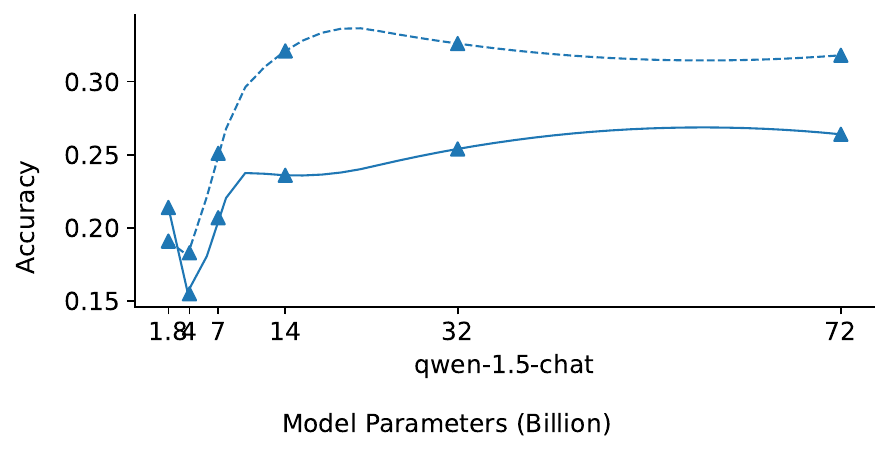}
	\caption{The accuracy of \textit{Mask} of Qwen-1.5-Chat models, the real and dashed lines represent the hard and easy split, respectively.}
	\label{fig: FRoG_scaling_qwen}
\end{figure}

\begin{figure*}[t!]
	\centering
	\includegraphics[width=1\textwidth]{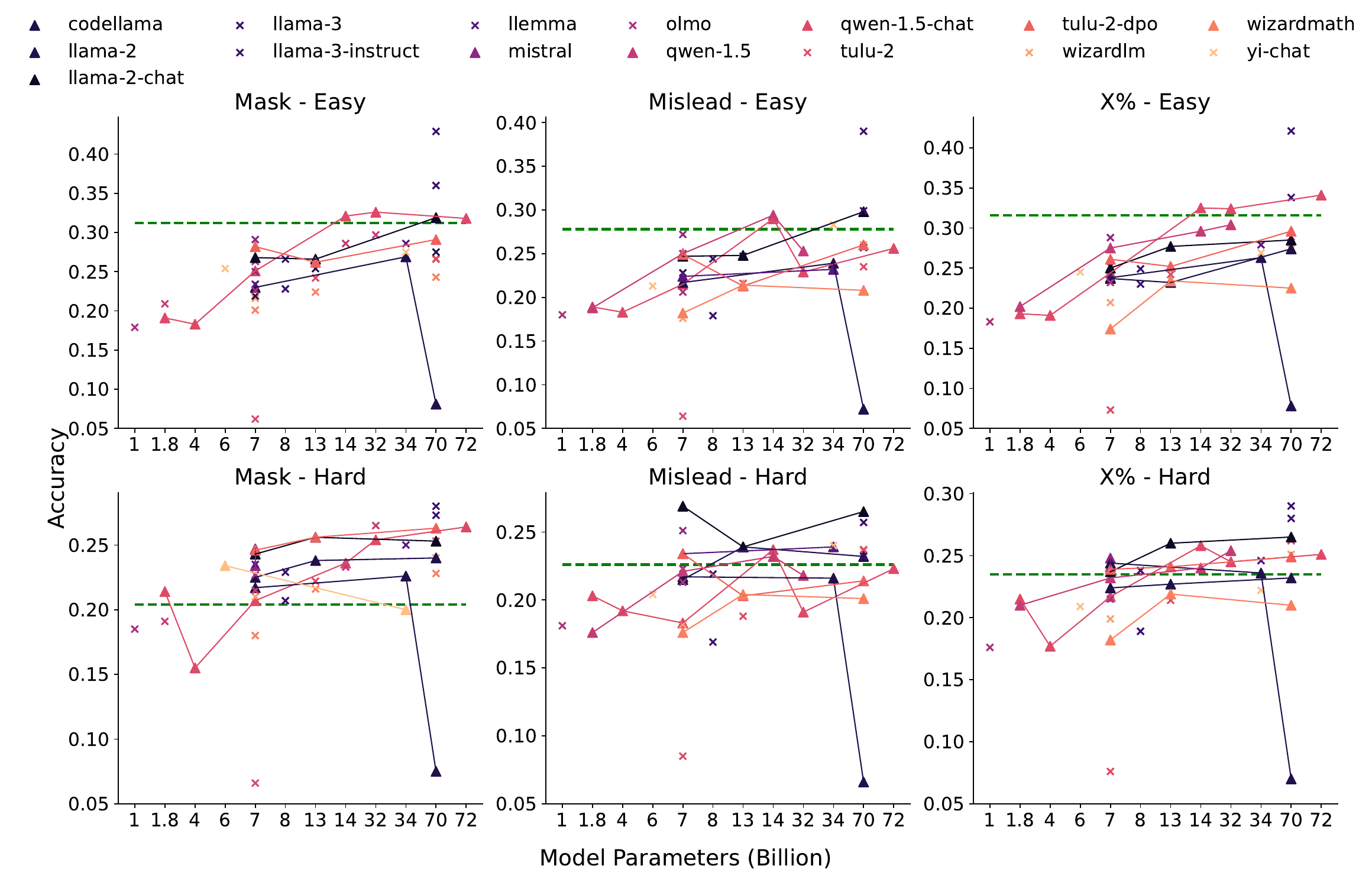}
	\caption{The performance of different LLMs on all \ourbenchmark tasks with different masking strategies and difficulties. The solid lines represent models that demonstrate inverse scaling phenomenon, and crossings represent the performance of other models. The green line represents the performance of GPT-3.5-turbo-1106. More than 50\% of the model families demonstrate the inverse scaling effect.}
	\label{fig: impact_inverse_scaling_all_models}
\end{figure*}

\subsubsection{Q2: Can the scaling law be observed in \ourbenchmarknospace?} 
Scaling law is introduced in \citet{kaplan2020scaling} to suggest the phenomenon that LLMs can achieve enhanced task performance by scaling up model sizes.
However, the scaling law does not hold universally. For example, the inverse scaling phenomenon can be observed when LLMs are instructed to choose which information can help to answer a question~\cite{mckenzie2023inverse}, which is similar to the design of \ourbenchmark tasks. Here, we list the performance of all models evaluated on \ourbenchmark in Figure~\ref{fig: impact_inverse_scaling_all_models}. Each line highlights an observed inverse scaling effect. It turns out that 8 out of the 15 model families evaluated demonstrate inverse scaling effect in \ourbenchmarknospace, crossing base models, continuous trained models and instruction-tuned models.

We further provide a case study of Qwen-1.5-Chat for its large number of open-resourced checkpoints and superior performance in mathematical reasoning. The results in Figure~\ref{fig: FRoG_scaling_qwen} show that the performance gap between \benchmarkeasy \Hquad and \benchmarkhard \Hquad increases starting Qwen-1.5-4B-Chat, and the performance becomes saturated with a model of 7 billion parameters or larger. Moreover, the performance gain of scaling model parameters diminishes after 14 billion model parameters. Inverse scaling happens on the side of models smaller than 7 billion and larger than 32 billion parameters. Models with fewer than 14 billion parameters are very unstable, displaying poor performance (below 25\%) and convoluted accuracy. Notably, the 4B model exhibits the poorest performance of all the models in \benchmarkeasy \Hquad and \benchmarkhard. Models with over 14 billion parameters, however, attain comparable performance in \benchmarkeasy.

\begin{figure*}[t!]
	\centering
        \includegraphics[width=0.9\textwidth]{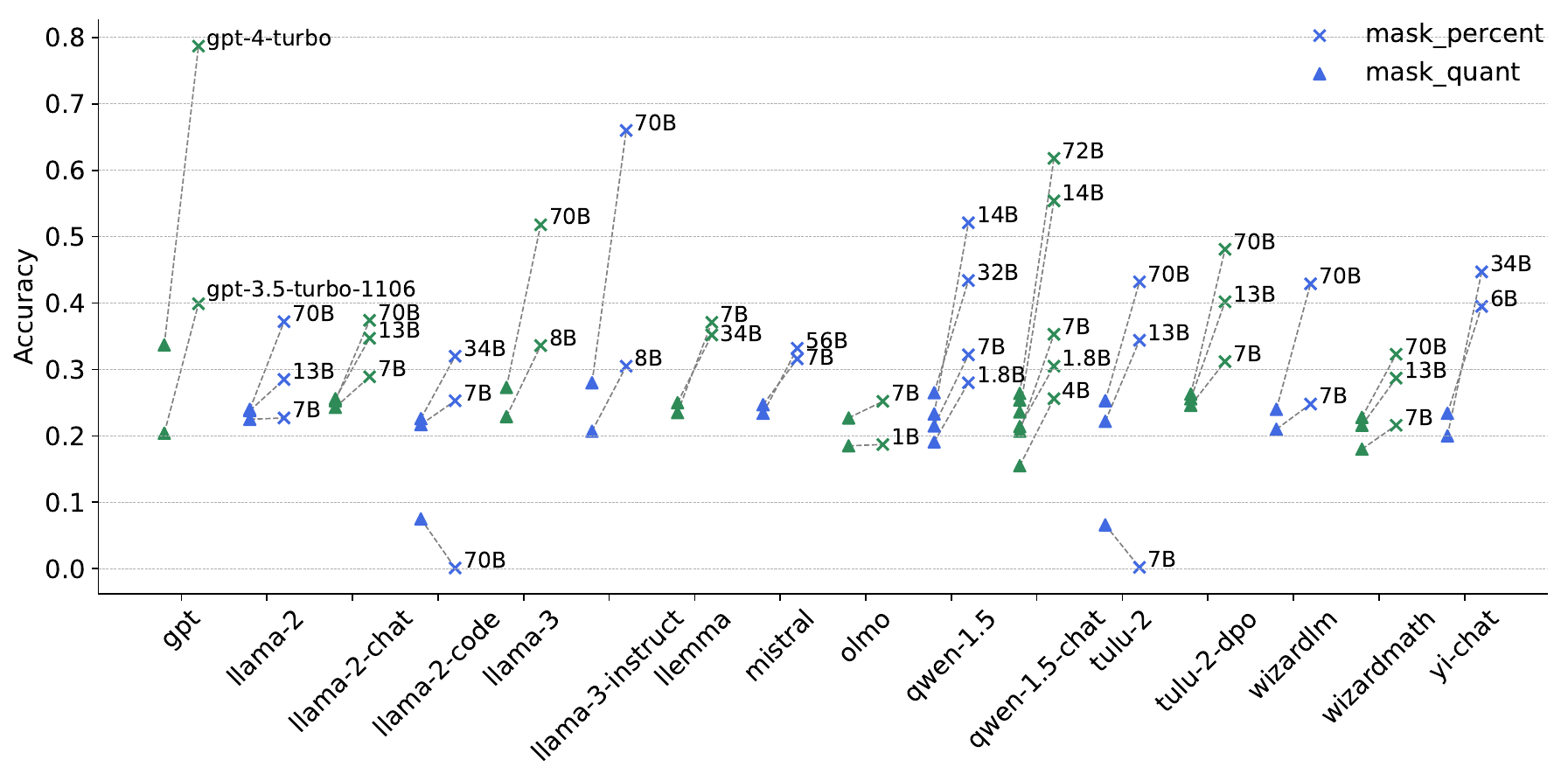}
	\caption{The performance comparison between the \textit{mask\_quant} and \textit{mask\_percent} on \benchmarkhard. Each dashed line connects the performance of the same model. LLMs with larger model sizes are more likely to receive larger performance degrade from \textit{mask\_percent} to \textit{mask\_quant} in \ourbenchmarknospace.}
	\label{fig: mask_percent_cmp}
\end{figure*}

\subsubsection{Q3: Is strong mathematical reasoning ability transferrable on \ourbenchmarknospace?} 
Mathematical reasoning has become a key signal for reasoning capabilities of LLMs~\cite{huang2024key}. To proxy the fuzzy reasoning ability under the precise setting, we design a \textit{mask\_percent} baseline that substitutes the misleading choices into their corresponding mean average strengths in QuRe and the correct choice being the exact value of the target percentage mention. Therefore, the original \textit{Mask} task (denoted as \textit{mask\_quant}) is transformed into figuring out the hidden percentage information in precise reasoning, with the predictions still reflect the preference over the quantifiers.

The results are demonstrated in Figure~\ref{fig: mask_percent_cmp}, where GPT-4-turbo and Llama-3-Instruct are the best performed commercial or open-resourced model in both \textit{mask\_percent} and \textit{mask\_quant}. CodeLlama-70B and Tulu-2-7B achieve minimal accuracy in both precise reasoning (nearly 0\%) and fuzzy reasoning (lower than 10\%) tasks, which attributes to their programming or web content style outputs. The accuracy of \textit{mask\_percent} of a model is significantly higher than the \textit{mask\_quant} alternatives in most of the cases. And the advances of \textit{mask\_percent} by scaling up model parameters are hard to transfer to the \textit{mask\_quant} task in \ourbenchmarknospace. Take Tulu-2-DPO models for instance, even though the \textit{mask\_percent} performance improves by scaling the model size, the performance of those models drops significantly and the scaling effect shrinks drastically when shifted to the \textit{mask\_quant} task. Specifically, we observe that models with a larger number of parameters are more likely to receive a larger accuracy drop when shifted from \textit{mask\_percent} to \textit{mask\_quant}. The average accuracy drop of models smaller than 10 billion parameters is 6.8\% compared to 16.9\% of models of larger sizes, indicating that successfully solving a reverse mathematical (precise) reasoning problem does not necessarily reflect equivalent competences in solving \ourbenchmark problems with fuzziness introduced by GQs. 

\begin{table*}[ht!]
    \small
    \centering
    \begin{tabular}{p{15cm}}
    \toprule        
        Ex1: Kelly has had 3 pay cuts in her salary in the past 6 months. If the first pay cut was 8\%, the second pay cut was 14\%, and the third was [MASK]. What will be the percentage decrease, if the salary is decreased in a single shot? If the answer to the question is 35.12\%, then please select the quantifier that is closest to the meaning of [MASK] from the following choices. [18\%]\\
        A. few B. some C. tiny amount \underline{D. small amount}\\
        \midrule
        Suppose that Kelly’s initial salary is 100 units for simplicity. After the first pay cut of 8\%, her salary becomes 92 units. The second pay cut of 14\% reduces her salary further to 92 * (1 - 0.14) = 79.12 units. Let the third pay cut be x\%. Therefore, her salary becomes 79.12 * (1 - x\%) after the third cut.
        
        We need to find x such that the overall reduction is equivalent to a single reduction of 35.12\%. This means the final salary should be 100 * (1 - 0.3512) = 64.88 units.
        
        Setting up the equation: 79.12 * (1 - x/100) = 64.88.
        
        Solving for x: (1 - x/100) = 64.88 / 79.12 $\fallingdotseq$ 0.8202, hence x/100 = 1 - 0.8202 = 0.1798 or x = 17.98\%.
        
        \quantfuzzyreason{Knowing the value of x, we can compare it to the options given:}
        
        \quantfuzzyreason{\quad - A few (typically refers to a small number, but greater than `tiny amount')}
        
        \quantfuzzyreason{\quad - Some (could imply a moderate to considerable amount, usually more ambiguous)}
        
        \quantfuzzyreason{\quad - Tiny amount (implies very small, likely less than our calculated 17.98\%)}
        
        \quantfuzzyreason{\quad - Small amount (suggests a non-trivial but not large percentage)}
        
        Between these choices, 'small amount' best describes a reduction of 17.98\%, as it's significant but not very large or tiny.
        Therefore, \textbf{the best choice is D. small amount.} \\
        \midrule
        Ex2: The average price of a decoration piece has increased over the years. If from 1990 to 1996, the price of the decoration piece increased by [MASK] and from 1996 to 2001 it increased by 20\%, what is the price of the decoration piece in 2001 if the price in 1990 was \$11500? If the answer to the question is \$15594, then please select the quantifier that is closest to the meaning of [MASK] from the following choices. [13\%] \\
        A. some B. tiny amount C. few \underline{D. small amount} \\
        \midrule
        First, we need to calculate the price of the decoration piece in 1996 based on an unknown percentage increase from 1990. Let the percentage increase from 1990 to 1996 be x\%.

        If the price in 1990 was \$11500, then the price in 1996 after x\% increase would be \$11500 * (1 + x/100).
        
        From 1996 to 2001, the price increased further by 20\%, so the price in 2001 would be:
        \$11500 * (1 + x/100) * 1.2
        
        The final value is given as \$15594.
        So we have: 
        \$11500 * (1 + x/100) * 1.2 = \$15594
        
        Solving for x:
        (1 + x/100) = \$15594 / (\$11500 * 1.2); 
        (1 + x/100) = \$15594 / \$13800;
        (1 + x/100) $\fallingdotseq$ 1.13;
        
        x/100 = 0.13
        x = 13\%
        
        So, the price increased by 13\% from 1990 to 1996.
        
        Given the choices of 'some,' 'tiny amount,' 'few,' and 'small amount,' a 13\% increase is not negligible but is not very large either. Among the provided options, 'small amount' seems to be the closest match to a 13\% increase.
        
        Therefore, the answer is \textbf{D. small amount}. \\
    \bottomrule
    \end{tabular}
    \caption{Sampled results in \benchmarkhard. The target percentage mention lies in the brackets, the correct answer is underscored and the prediction is bolded. The \quantfuzzyreason{explicit quantifier estimation stage} is highlighted.}
    \label{tab: FRoG examples fuzzy mechanisms}
\end{table*}

\subsection{Case Study}
\label{sec: case study}
Lastly, we are interested in the working mechanism that LLMs pose in conducting fuzzy reasoning in \ourbenchmarknospace. We randomly sample 50 results from GPT-4-turbo in \benchmarkhard, and demonstrate examples in Table~\ref{tab: FRoG examples fuzzy mechanisms}. The mechanism of LLMs to solve fuzzy reasoning can be summarized as follows: firstly, the LLMs attempt to solve the target percentage mention in three major patterns:\\
(1) Precisely solving the target percentage mention. \\
(2) Reaching a percentage value that is close to the target percentage mention. \\
(3) Incomplete solution (e.g. unsolved equations). \\
After solving the target percentage mention, the LLMs select their preferred quantifiers through a quantifier estimation stage, which can be implicit. Such implicit reasoning procedure is also studied in precise math reasoning~\cite{deng2023implicit, deng2024explicit}.

For example in Ex1 of Table~\ref{tab: FRoG examples fuzzy mechanisms}, the model first solves the target percentage mention with satisfying precision (17.98\%), and then conduct a quantifier estimation stage before selecting the quantifier. Note that this stage is not always explicit in selecting the correct quantifier (e.g. Ex2). We include more examples in Appendix~\ref{sec: more example in fuzzy reasoning mechansims}.

In general, the target percentage mention can be solved with satisfying precision (within 1\% error) 52\% of the time (still lower than \textit{mask\_percent} where the target percentage value is given as a candidate choice), with only 61\% of them lead to the final correctness of the problems. It indicates the complexity of conducting fuzzy reasoning even with the target percentage value given.  

Moreover, 15 of the 50 predictions show explicit quantifier estimation by comparing or estimating the strengths of quantifiers, but only 5 of them lead to the correct quantifier predictions, meaning that precisely solving the target percentage value is insufficient to conduct fuzzy reasoning. Meanwhile, there are 15 cases where models select the correct quantifier without explicit quantifier estimations, indicating that models may rely on implicit mechanisms in conducting fuzzy reasoning.

Note that even though models within the same model family but with different number of parameters can figure out the target percentage mention correctly or close enough, their interpretation of quantifiers can make a difference in the final prediction, we refer to Appendix~\ref{sec: quantifier understanding across scale} for examples.

\section{Conclusion}
The fuzzy reasoning ability is an under-explored direction of the reasoning ability of LLMs. To measure the fuzzy reasoning ability of LLMs, we collect a fuzzy reasoning benchmark \ourbenchmark that is based on generalized quantifiers. The experimental results show that fuzzy reasoning remains challenging for current LLMs, and an inverse scaling effect is observed on the performance of \ourbenchmarknospace. Besides, prevailing reasoning enhancement approaches including continuous pretraining, instruction tuning and general alignment may not stay effective on fuzzy reasoning of \ourbenchmarknospace. Lastly, LLMs can demonstrate diverse behaviors in fuzzy reasoning. 

\section*{Limitations}
In this work, we collect a fuzzy reasoning dataset \ourbenchmarknospace ~to evaluate the fuzzy reasoning abilities of several existing LLMs. We are aware that even though the problems in \ourbenchmarknospace ~originate from real-world math word problems, the new question created may not naturally occur, and the designed masking-based reasoning protocol is not identical to the real-world reasoning procedure where the vague information is processed directly. We also note that GQ-based fuzzy reasoning is only a subset of the entire family of natural language fuzzy reasoning, and the scope of GQs is broader than the ones being studied in this work.

\bibliography{anthology,custom}

\begin{thebibliography}{67}
\providecommand{\natexlab}[1]{#1}

\bibitem[{AI(2024)}]{llama-3}
Meta AI. 2024.
\newblock Introducing meta llama 3: The most capable openly available llm to date.

\bibitem[{Amini et~al.(2019)Amini, Gabriel, Lin, Koncel-Kedziorski, Choi, and Hajishirzi}]{amini-etal-2019-mathqa}
Aida Amini, Saadia Gabriel, Shanchuan Lin, Rik Koncel-Kedziorski, Yejin Choi, and Hannaneh Hajishirzi. 2019.
\newblock \href {https://doi.org/10.18653/v1/N19-1245} {{M}ath{QA}: Towards interpretable math word problem solving with operation-based formalisms}.
\newblock In \emph{Proceedings of the 2019 Conference of the North {A}merican Chapter of the Association for Computational Linguistics: Human Language Technologies, Volume 1 (Long and Short Papers)}, pages 2357--2367, Minneapolis, Minnesota. Association for Computational Linguistics.

\bibitem[{An et~al.(2023)An, Ma, Lin, Zheng, Lou, and Chen}]{an2023learning}
Shengnan An, Zexiong Ma, Zeqi Lin, Nanning Zheng, Jian-Guang Lou, and Weizhu Chen. 2023.
\newblock Learning from mistakes makes llm better reasoner.
\newblock \emph{arXiv preprint arXiv:2310.20689}.

\bibitem[{Azerbayev et~al.(2024)Azerbayev, Schoelkopf, Paster, Santos, McAleer, Jiang, Deng, Biderman, and Welleck}]{azerbayev2024llemma}
Zhangir Azerbayev, Hailey Schoelkopf, Keiran Paster, Marco~Dos Santos, Stephen~Marcus McAleer, Albert~Q. Jiang, Jia Deng, Stella Biderman, and Sean Welleck. 2024.
\newblock \href {https://openreview.net/forum?id=4WnqRR915j} {Llemma: An open language model for mathematics}.
\newblock In \emph{The Twelfth International Conference on Learning Representations}.

\bibitem[{Bai et~al.(2023)Bai, Bai, Chu, Cui, Dang, Deng, Fan, Ge, Han, Huang, Hui, Ji, Li, Lin, Lin, Liu, Liu, Lu, Lu, Ma, Men, Ren, Ren, Tan, Tan, Tu, Wang, Wang, Wang, Wu, Xu, Xu, Yang, Yang, Yang, Yang, Yao, Yu, Yuan, Yuan, Zhang, Zhang, Zhang, Zhang, Zhou, Zhou, Zhou, and Zhu}]{qwen}
Jinze Bai, Shuai Bai, Yunfei Chu, Zeyu Cui, Kai Dang, Xiaodong Deng, Yang Fan, Wenbin Ge, Yu~Han, Fei Huang, Binyuan Hui, Luo Ji, Mei Li, Junyang Lin, Runji Lin, Dayiheng Liu, Gao Liu, Chengqiang Lu, Keming Lu, Jianxin Ma, Rui Men, Xingzhang Ren, Xuancheng Ren, Chuanqi Tan, Sinan Tan, Jianhong Tu, Peng Wang, Shijie Wang, Wei Wang, Shengguang Wu, Benfeng Xu, Jin Xu, An~Yang, Hao Yang, Jian Yang, Shusheng Yang, Yang Yao, Bowen Yu, Hongyi Yuan, Zheng Yuan, Jianwei Zhang, Xingxuan Zhang, Yichang Zhang, Zhenru Zhang, Chang Zhou, Jingren Zhou, Xiaohuan Zhou, and Tianhang Zhu. 2023.
\newblock Qwen technical report.
\newblock \emph{arXiv preprint arXiv:2309.16609}.

\bibitem[{Bowen et~al.(2024)Bowen, S{\ae}tre, and Miyao}]{bowen-etal-2024-comprehensive}
Chen Bowen, Rune S{\ae}tre, and Yusuke Miyao. 2024.
\newblock \href {https://aclanthology.org/2024.findings-eacl.22} {A comprehensive evaluation of inductive reasoning capabilities and problem solving in large language models}.
\newblock In \emph{Findings of the Association for Computational Linguistics: EACL 2024}, pages 323--339, St. Julian{'}s, Malta. Association for Computational Linguistics.

\bibitem[{Brown et~al.(2020)Brown, Mann, Ryder, Subbiah, Kaplan, Dhariwal, Neelakantan, Shyam, Sastry, Askell, Agarwal, Herbert-Voss, Krueger, Henighan, Child, Ramesh, Ziegler, Wu, Winter, Hesse, Chen, Sigler, Litwin, Gray, Chess, Clark, Berner, McCandlish, Radford, Sutskever, and Amodei}]{NEURIPS2020_GPT3}
Tom Brown, Benjamin Mann, Nick Ryder, Melanie Subbiah, Jared~D Kaplan, Prafulla Dhariwal, Arvind Neelakantan, Pranav Shyam, Girish Sastry, Amanda Askell, Sandhini Agarwal, Ariel Herbert-Voss, Gretchen Krueger, Tom Henighan, Rewon Child, Aditya Ramesh, Daniel Ziegler, Jeffrey Wu, Clemens Winter, Chris Hesse, Mark Chen, Eric Sigler, Mateusz Litwin, Scott Gray, Benjamin Chess, Jack Clark, Christopher Berner, Sam McCandlish, Alec Radford, Ilya Sutskever, and Dario Amodei. 2020.
\newblock \href {https://proceedings.neurips.cc/paper_files/paper/2020/file/1457c0d6bfcb4967418bfb8ac142f64a-Paper.pdf} {Language models are few-shot learners}.
\newblock In \emph{Advances in Neural Information Processing Systems}, volume~33, pages 1877--1901. Curran Associates, Inc.

\bibitem[{Chen et~al.(2023{\natexlab{a}})Chen, Pan, Yu, Song, Wang, Yu, and Chen}]{chen2023skills}
Jiaao Chen, Xiaoman Pan, Dian Yu, Kaiqiang Song, Xiaoyang Wang, Dong Yu, and Jianshu Chen. 2023{\natexlab{a}}.
\newblock Skills-in-context prompting: Unlocking compositionality in large language models.
\newblock \emph{arXiv preprint arXiv:2308.00304}.

\bibitem[{Chen et~al.(2023{\natexlab{b}})Chen, Ma, Wang, and Cohen}]{chen2023program}
Wenhu Chen, Xueguang Ma, Xinyi Wang, and William~W. Cohen. 2023{\natexlab{b}}.
\newblock \href {https://openreview.net/forum?id=YfZ4ZPt8zd} {Program of thoughts prompting: Disentangling computation from reasoning for numerical reasoning tasks}.
\newblock \emph{Transactions on Machine Learning Research}.

\bibitem[{Chia et~al.(2023)Chia, Chen, Tuan, Poria, and Bing}]{chia2023contrastive}
Yew~Ken Chia, Guizhen Chen, Luu~Anh Tuan, Soujanya Poria, and Lidong Bing. 2023.
\newblock Contrastive chain-of-thought prompting.
\newblock \emph{arXiv preprint arXiv:2311.09277}.

\bibitem[{Cobbe et~al.(2021)Cobbe, Kosaraju, Bavarian, Chen, Jun, Kaiser, Plappert, Tworek, Hilton, Nakano, Hesse, and Schulman}]{cobbe2021gsm8k}
Karl Cobbe, Vineet Kosaraju, Mohammad Bavarian, Mark Chen, Heewoo Jun, Lukasz Kaiser, Matthias Plappert, Jerry Tworek, Jacob Hilton, Reiichiro Nakano, Christopher Hesse, and John Schulman. 2021.
\newblock Training verifiers to solve math word problems.
\newblock \emph{arXiv preprint arXiv:2110.14168}.

\bibitem[{Cui et~al.(2022)Cui, Hershcovich, and S{\o}gaard}]{cui-etal-2022-generalized-quantifiers}
Ruixiang Cui, Daniel Hershcovich, and Anders S{\o}gaard. 2022.
\newblock \href {https://doi.org/10.18653/v1/2022.naacl-main.359} {Generalized quantifiers as a source of error in multilingual {NLU} benchmarks}.
\newblock In \emph{Proceedings of the 2022 Conference of the North American Chapter of the Association for Computational Linguistics: Human Language Technologies}, pages 4875--4893, Seattle, United States. Association for Computational Linguistics.

\bibitem[{DeepSeek-AI et~al.(2024)DeepSeek-AI, :, Bi, Chen, Chen, Chen, Dai, Deng, Ding, Dong, Du, Fu, Gao, Gao, Gao, Ge, Guan, Guo, Guo, Hao, Hao, He, Hu, Huang, Li, Li, Li, Li, Li, Liang, Lin, Liu, Liu, Liu, Liu, Liu, Liu, Lu, Lu, Luo, Ma, Nie, Pei, Piao, Qiu, Qu, Ren, Ren, Ruan, Sha, Shao, Song, Su, Sun, Sun, Tang, Wang, Wang, Wang, Wang, Wang, Wu, Wu, Xie, Xie, Xie, Xiong, Xu, Xu, Xu, Yang, You, Yu, Yu, Zhang, Zhang, Zhang, Zhang, Zhang, Zhang, Zhang, Zhang, Zhao, Zhao, Zhou, Zhou, Zhu, and Zou}]{deepseekai2024deepseek}
DeepSeek-AI, :, Xiao Bi, Deli Chen, Guanting Chen, Shanhuang Chen, Damai Dai, Chengqi Deng, Honghui Ding, Kai Dong, Qiushi Du, Zhe Fu, Huazuo Gao, Kaige Gao, Wenjun Gao, Ruiqi Ge, Kang Guan, Daya Guo, Jianzhong Guo, Guangbo Hao, Zhewen Hao, Ying He, Wenjie Hu, Panpan Huang, Erhang Li, Guowei Li, Jiashi Li, Yao Li, Y.~K. Li, Wenfeng Liang, Fangyun Lin, A.~X. Liu, Bo~Liu, Wen Liu, Xiaodong Liu, Xin Liu, Yiyuan Liu, Haoyu Lu, Shanghao Lu, Fuli Luo, Shirong Ma, Xiaotao Nie, Tian Pei, Yishi Piao, Junjie Qiu, Hui Qu, Tongzheng Ren, Zehui Ren, Chong Ruan, Zhangli Sha, Zhihong Shao, Junxiao Song, Xuecheng Su, Jingxiang Sun, Yaofeng Sun, Minghui Tang, Bingxuan Wang, Peiyi Wang, Shiyu Wang, Yaohui Wang, Yongji Wang, Tong Wu, Y.~Wu, Xin Xie, Zhenda Xie, Ziwei Xie, Yiliang Xiong, Hanwei Xu, R.~X. Xu, Yanhong Xu, Dejian Yang, Yuxiang You, Shuiping Yu, Xingkai Yu, B.~Zhang, Haowei Zhang, Lecong Zhang, Liyue Zhang, Mingchuan Zhang, Minghua Zhang, Wentao Zhang, Yichao Zhang, Chenggang Zhao, Yao Zhao, Shangyan Zhou, Shunfeng
  Zhou, Qihao Zhu, and Yuheng Zou. 2024.
\newblock \href {https://arxiv.org/abs/2401.02954} {Deepseek llm: Scaling open-source language models with longtermism}.
\newblock \emph{Preprint}, arXiv:2401.02954.

\bibitem[{Deng et~al.(2024)Deng, Choi, and Shieber}]{deng2024explicit}
Yuntian Deng, Yejin Choi, and Stuart Shieber. 2024.
\newblock \href {https://arxiv.org/abs/2405.14838} {From explicit cot to implicit cot: Learning to internalize cot step by step}.
\newblock \emph{Preprint}, arXiv:2405.14838.

\bibitem[{Deng et~al.(2023)Deng, Prasad, Fernandez, Smolensky, Chaudhary, and Shieber}]{deng2023implicit}
Yuntian Deng, Kiran Prasad, Roland Fernandez, Paul Smolensky, Vishrav Chaudhary, and Stuart Shieber. 2023.
\newblock \href {https://arxiv.org/abs/2311.01460} {Implicit chain of thought reasoning via knowledge distillation}.
\newblock \emph{Preprint}, arXiv:2311.01460.

\bibitem[{Faghihi et~al.(2020)Faghihi, Robert, Poirier, and Barkaoui}]{faghihi2020association}
Usef Faghihi, Serge Robert, Pierre Poirier, and Youssef Barkaoui. 2020.
\newblock From association to reasoning, an alternative to pearls’ causal reasoning.
\newblock In \emph{The Thirty-Third International Flairs Conference}.

\bibitem[{Gao et~al.(2023)Gao, Madaan, Zhou, Alon, Liu, Yang, Callan, and Neubig}]{gao2023pal}
Luyu Gao, Aman Madaan, Shuyan Zhou, Uri Alon, Pengfei Liu, Yiming Yang, Jamie Callan, and Graham Neubig. 2023.
\newblock Pal: Program-aided language models.
\newblock In \emph{International Conference on Machine Learning}, pages 10764--10799. PMLR.

\bibitem[{Gou et~al.(2023{\natexlab{a}})Gou, Shao, Gong, Shen, Yang, Duan, and Chen}]{gou2023critic}
Zhibin Gou, Zhihong Shao, Yeyun Gong, Yelong Shen, Yujiu Yang, Nan Duan, and Weizhu Chen. 2023{\natexlab{a}}.
\newblock Critic: Large language models can self-correct with tool-interactive critiquing.
\newblock \emph{arXiv preprint arXiv:2305.11738}.

\bibitem[{Gou et~al.(2023{\natexlab{b}})Gou, Shao, Gong, Yang, Huang, Duan, Chen et~al.}]{gou2023tora}
Zhibin Gou, Zhihong Shao, Yeyun Gong, Yujiu Yang, Minlie Huang, Nan Duan, Weizhu Chen, et~al. 2023{\natexlab{b}}.
\newblock Tora: A tool-integrated reasoning agent for mathematical problem solving.
\newblock \emph{arXiv preprint arXiv:2309.17452}.

\bibitem[{Hendrycks et~al.(2021)Hendrycks, Burns, Kadavath, Arora, Basart, Tang, Song, and Steinhardt}]{hendrycksmath2021}
Dan Hendrycks, Collin Burns, Saurav Kadavath, Akul Arora, Steven Basart, Eric Tang, Dawn Song, and Jacob Steinhardt. 2021.
\newblock Measuring mathematical problem solving with the math dataset.
\newblock \emph{NeurIPS}.

\bibitem[{Hu et~al.(2024)Hu, Chen, and Ponti}]{hu2024fine}
Hanxu Hu, Pinzhen Chen, and Edoardo~M Ponti. 2024.
\newblock Fine-tuning large language models with sequential instructions.
\newblock \emph{arXiv preprint arXiv:2403.07794}.

\bibitem[{Huang and Chang(2022)}]{huang2022towards}
Jie Huang and Kevin Chen-Chuan Chang. 2022.
\newblock Towards reasoning in large language models: A survey.
\newblock \emph{arXiv preprint arXiv:2212.10403}.

\bibitem[{Huang et~al.(2024{\natexlab{a}})Huang, Liu, Gong, Gou, Shen, Duan, and Chen}]{huang2024key}
Yiming Huang, Xiao Liu, Yeyun Gong, Zhibin Gou, Yelong Shen, Nan Duan, and Weizhu Chen. 2024{\natexlab{a}}.
\newblock Key-point-driven data synthesis with its enhancement on mathematical reasoning.
\newblock \emph{arXiv preprint arXiv:2403.02333}.

\bibitem[{Huang et~al.(2024{\natexlab{b}})Huang, Lin, Liu, Cao, Xin, Wang, Li, Song, and Liang}]{huang2024mustard}
Yinya Huang, Xiaohan Lin, Zhengying Liu, Qingxing Cao, Huajian Xin, Haiming Wang, Zhenguo Li, Linqi Song, and Xiaodan Liang. 2024{\natexlab{b}}.
\newblock Mustard: Mastering uniform synthesis of theorem and proof data.
\newblock \emph{arXiv preprint arXiv:2402.08957}.

\bibitem[{Ivison* et~al.(2023)Ivison*, Wang*, Pyatkin, Lambert, Peters, Dasigi, Jang, Wadden, Smith, Beltagy, and Hajishirzi}]{ivison2023camels}
Hamish Ivison*, Yizhong Wang*, Valentina Pyatkin, Nathan Lambert, Matthew Peters, Pradeep Dasigi, Joel Jang, David Wadden, Noah~A. Smith, Iz~Beltagy, and Hannaneh Hajishirzi. 2023.
\newblock \href {https://arxiv.org/abs/2311.10702} {Camels in a changing climate: Enhancing lm adaptation with tulu 2}.
\newblock \emph{arXiv preprint}.

\bibitem[{Jiang et~al.(2023)Jiang, Sablayrolles, Mensch, Bamford, Chaplot, Casas, Bressand, Lengyel, Lample, Saulnier et~al.}]{jiang2023mistral}
Albert~Q Jiang, Alexandre Sablayrolles, Arthur Mensch, Chris Bamford, Devendra~Singh Chaplot, Diego de~las Casas, Florian Bressand, Gianna Lengyel, Guillaume Lample, Lucile Saulnier, et~al. 2023.
\newblock Mistral 7b.
\newblock \emph{arXiv preprint arXiv:2310.06825}.

\bibitem[{Joshi et~al.(2020)Joshi, Aditya, Sathe, and Choudhury}]{joshi-etal-2020-taxinli}
Pratik Joshi, Somak Aditya, Aalok Sathe, and Monojit Choudhury. 2020.
\newblock \href {https://doi.org/10.18653/v1/2020.conll-1.4} {{T}axi{NLI}: Taking a ride up the {NLU} hill}.
\newblock In \emph{Proceedings of the 24th Conference on Computational Natural Language Learning}, pages 41--55, Online. Association for Computational Linguistics.

\bibitem[{Kaplan et~al.(2020)Kaplan, McCandlish, Henighan, Brown, Chess, Child, Gray, Radford, Wu, and Amodei}]{kaplan2020scaling}
Jared Kaplan, Sam McCandlish, Tom Henighan, Tom~B. Brown, Benjamin Chess, Rewon Child, Scott Gray, Alec Radford, Jeffrey Wu, and Dario Amodei. 2020.
\newblock \href {https://arxiv.org/abs/2001.08361} {Scaling laws for neural language models}.
\newblock \emph{Preprint}, arXiv:2001.08361.

\bibitem[{Kapustin and Kapustin(2019)}]{kapustin-kapustin-2019-modeling}
Pavlo Kapustin and Michael Kapustin. 2019.
\newblock \href {https://doi.org/10.18653/v1/W19-0604} {Modeling language constructs with fuzzy sets: some approaches, examples and interpretations}.
\newblock In \emph{Proceedings of the 13th International Conference on Computational Semantics - Student Papers}, pages 24--33, Gothenburg, Sweden. Association for Computational Linguistics.

\bibitem[{Lee(2004)}]{lee2004first}
Kwang~Hyung Lee. 2004.
\newblock \emph{First course on fuzzy theory and applications}, volume~27.
\newblock Springer Science \& Business Media.

\bibitem[{Lewkowycz et~al.(2022)Lewkowycz, Andreassen, Dohan, Dyer, Michalewski, Ramasesh, Slone, Anil, Schlag, Gutman-Solo, Wu, Neyshabur, Gur-Ari, and Misra}]{lewkowycz2022solving}
Aitor Lewkowycz, Anders~Johan Andreassen, David Dohan, Ethan Dyer, Henryk Michalewski, Vinay~Venkatesh Ramasesh, Ambrose Slone, Cem Anil, Imanol Schlag, Theo Gutman-Solo, Yuhuai Wu, Behnam Neyshabur, Guy Gur-Ari, and Vedant Misra. 2022.
\newblock \href {https://openreview.net/forum?id=IFXTZERXdM7} {Solving quantitative reasoning problems with language models}.
\newblock In \emph{Advances in Neural Information Processing Systems}.

\bibitem[{Li et~al.(2024)Li, Dong, Tang, Wang, Zhang, Huang, Huang, Huang, Huang, Zhang et~al.}]{li2024synthetic}
Haoran Li, Qingxiu Dong, Zhengyang Tang, Chaojun Wang, Xingxing Zhang, Haoyang Huang, Shaohan Huang, Xiaolong Huang, Zeqiang Huang, Dongdong Zhang, et~al. 2024.
\newblock Synthetic data (almost) from scratch: Generalized instruction tuning for language models.
\newblock \emph{arXiv preprint arXiv:2402.13064}.

\bibitem[{Li et~al.(2023)Li, Menon, Ghosh, and Srivastava}]{li-etal-2023-pragmatic}
Yiyuan Li, Rakesh Menon, Sayan Ghosh, and Shashank Srivastava. 2023.
\newblock \href {https://doi.org/10.18653/v1/2023.emnlp-main.38} {Pragmatic reasoning unlocks quantifier semantics for foundation models}.
\newblock In \emph{Proceedings of the 2023 Conference on Empirical Methods in Natural Language Processing}, pages 573--591, Singapore. Association for Computational Linguistics.

\bibitem[{Luo et~al.(2023)Luo, Sun, Xu, Zhao, Lou, Tao, Geng, Lin, Chen, and Zhang}]{luo2023wizardmath}
Haipeng Luo, Qingfeng Sun, Can Xu, Pu~Zhao, Jianguang Lou, Chongyang Tao, Xiubo Geng, Qingwei Lin, Shifeng Chen, and Dongmei Zhang. 2023.
\newblock \href {https://arxiv.org/abs/2308.09583} {Wizardmath: Empowering mathematical reasoning for large language models via reinforced evol-instruct}.
\newblock \emph{Preprint}, arXiv:2308.09583.

\bibitem[{McKenzie et~al.(2023)McKenzie, Lyzhov, Pieler, Parrish, Mueller, Prabhu, McLean, Shen, Cavanagh, Gritsevskiy, Kauffman, Kirtland, Zhou, Zhang, Huang, Wurgaft, Weiss, Ross, Recchia, Liu, Liu, Tseng, Korbak, Kim, Bowman, and Perez}]{mckenzie2023inverse}
Ian~R. McKenzie, Alexander Lyzhov, Michael~Martin Pieler, Alicia Parrish, Aaron Mueller, Ameya Prabhu, Euan McLean, Xudong Shen, Joe Cavanagh, Andrew~George Gritsevskiy, Derik Kauffman, Aaron~T. Kirtland, Zhengping Zhou, Yuhui Zhang, Sicong Huang, Daniel Wurgaft, Max Weiss, Alexis Ross, Gabriel Recchia, Alisa Liu, Jiacheng Liu, Tom Tseng, Tomasz Korbak, Najoung Kim, Samuel~R. Bowman, and Ethan Perez. 2023.
\newblock \href {https://openreview.net/forum?id=DwgRm72GQF} {Inverse scaling: When bigger isn't better}.
\newblock \emph{Transactions on Machine Learning Research}.
\newblock Featured Certification.

\bibitem[{Mishra et~al.(2022)Mishra, Finlayson, Lu, Tang, Welleck, Baral, Rajpurohit, Tafjord, Sabharwal, Clark et~al.}]{mishra2022lila}
Swaroop Mishra, Matthew Finlayson, Pan Lu, Leonard Tang, Sean Welleck, Chitta Baral, Tanmay Rajpurohit, Oyvind Tafjord, Ashish Sabharwal, Peter Clark, et~al. 2022.
\newblock Lila: A unified benchmark for mathematical reasoning.
\newblock \emph{arXiv preprint arXiv:2210.17517}.

\bibitem[{Mostowski(1957)}]{Mostowski1957}
Andrzej Mostowski. 1957.
\newblock \href {http://eudml.org/doc/213418} {On a generalization of quantifiers}.
\newblock \emph{Fundamenta Mathematicae}, 44(1):12--36.

\bibitem[{Nov{\'a}k(2015)}]{Novák2015}
Vil{\'e}m Nov{\'a}k. 2015.
\newblock \href {https://doi.org/10.1007/978-3-319-18750-1_8} {\emph{Fuzzy Natural Logic: Towards Mathematical Logic of Human Reasoning}}, pages 137--165.
\newblock Springer International Publishing, Cham.

\bibitem[{Ouyang et~al.(2024)Ouyang, Wu, Jiang, Almeida, Wainwright, Mishkin, Zhang, Agarwal, Slama, Ray, Schulman, Hilton, Kelton, Miller, Simens, Askell, Welinder, Christiano, Leike, and Lowe}]{instructgpt}
Long Ouyang, Jeff Wu, Xu~Jiang, Diogo Almeida, Carroll~L. Wainwright, Pamela Mishkin, Chong Zhang, Sandhini Agarwal, Katarina Slama, Alex Ray, John Schulman, Jacob Hilton, Fraser Kelton, Luke Miller, Maddie Simens, Amanda Askell, Peter Welinder, Paul Christiano, Jan Leike, and Ryan Lowe. 2024.
\newblock Training language models to follow instructions with human feedback.
\newblock In \emph{Proceedings of the 36th International Conference on Neural Information Processing Systems}, NIPS '22, Red Hook, NY, USA. Curran Associates Inc.

\bibitem[{Paster et~al.(2023)Paster, Santos, Azerbayev, and Ba}]{paster2023openwebmath}
Keiran Paster, Marco~Dos Santos, Zhangir Azerbayev, and Jimmy Ba. 2023.
\newblock Openwebmath: An open dataset of high-quality mathematical web text.
\newblock \emph{arXiv preprint arXiv:2310.06786}.

\bibitem[{Rafailov et~al.(2023)Rafailov, Sharma, Mitchell, Manning, Ermon, and Finn}]{rafailov2023direct}
Rafael Rafailov, Archit Sharma, Eric Mitchell, Christopher~D Manning, Stefano Ermon, and Chelsea Finn. 2023.
\newblock \href {https://openreview.net/forum?id=HPuSIXJaa9} {Direct preference optimization: Your language model is secretly a reward model}.
\newblock In \emph{Thirty-seventh Conference on Neural Information Processing Systems}.

\bibitem[{Rozière et~al.(2024)Rozière, Gehring, Gloeckle, Sootla, Gat, Tan, Adi, Liu, Sauvestre, Remez, Rapin, Kozhevnikov, Evtimov, Bitton, Bhatt, Ferrer, Grattafiori, Xiong, Défossez, Copet, Azhar, Touvron, Martin, Usunier, Scialom, and Synnaeve}]{rozière2024code}
Baptiste Rozière, Jonas Gehring, Fabian Gloeckle, Sten Sootla, Itai Gat, Xiaoqing~Ellen Tan, Yossi Adi, Jingyu Liu, Romain Sauvestre, Tal Remez, Jérémy Rapin, Artyom Kozhevnikov, Ivan Evtimov, Joanna Bitton, Manish Bhatt, Cristian~Canton Ferrer, Aaron Grattafiori, Wenhan Xiong, Alexandre Défossez, Jade Copet, Faisal Azhar, Hugo Touvron, Louis Martin, Nicolas Usunier, Thomas Scialom, and Gabriel Synnaeve. 2024.
\newblock \href {https://arxiv.org/abs/2308.12950} {Code llama: Open foundation models for code}.
\newblock \emph{Preprint}, arXiv:2308.12950.

\bibitem[{Saki and Faghihi(2022)}]{saki2022fundamental}
Amir Saki and Usef Faghihi. 2022.
\newblock \href {https://arxiv.org/abs/2205.15016} {A fundamental probabilistic fuzzy logic framework suitable for causal reasoning}.
\newblock \emph{Preprint}, arXiv:2205.15016.

\bibitem[{Suhr et~al.(2019)Suhr, Zhou, Zhang, Zhang, Bai, and Artzi}]{suhr-etal-2019-corpus}
Alane Suhr, Stephanie Zhou, Ally Zhang, Iris Zhang, Huajun Bai, and Yoav Artzi. 2019.
\newblock \href {https://doi.org/10.18653/v1/P19-1644} {A corpus for reasoning about natural language grounded in photographs}.
\newblock In \emph{Proceedings of the 57th Annual Meeting of the Association for Computational Linguistics}, pages 6418--6428, Florence, Italy. Association for Computational Linguistics.

\bibitem[{Tang et~al.(2024)Tang, Zhang, Wan, and Wei}]{tang2024mathscale}
Zhengyang Tang, Xingxing Zhang, Benyou Wan, and Furu Wei. 2024.
\newblock Mathscale: Scaling instruction tuning for mathematical reasoning.
\newblock \emph{arXiv preprint arXiv:2403.02884}.

\bibitem[{Taylor et~al.(2022)Taylor, Kardas, Cucurull, Scialom, Hartshorn, Saravia, Poulton, Kerkez, and Stojnic}]{taylor2022galactica}
Ross Taylor, Marcin Kardas, Guillem Cucurull, Thomas Scialom, Anthony Hartshorn, Elvis Saravia, Andrew Poulton, Viktor Kerkez, and Robert Stojnic. 2022.
\newblock \href {https://arxiv.org/abs/2211.09085} {Galactica: A large language model for science}.
\newblock \emph{Preprint}, arXiv:2211.09085.

\bibitem[{Touvron et~al.(2023)Touvron, Martin, Stone, Albert, Almahairi, Babaei, Bashlykov, Batra, Bhargava, Bhosale, Bikel, Blecher, Ferrer, Chen, Cucurull, Esiobu, Fernandes, Fu, Fu, Fuller, Gao, Goswami, Goyal, Hartshorn, Hosseini, Hou, Inan, Kardas, Kerkez, Khabsa, Kloumann, Korenev, Koura, Lachaux, Lavril, Lee, Liskovich, Lu, Mao, Martinet, Mihaylov, Mishra, Molybog, Nie, Poulton, Reizenstein, Rungta, Saladi, Schelten, Silva, Smith, Subramanian, Tan, Tang, Taylor, Williams, Kuan, Xu, Yan, Zarov, Zhang, Fan, Kambadur, Narang, Rodriguez, Stojnic, Edunov, and Scialom}]{touvron2023llama}
Hugo Touvron, Louis Martin, Kevin Stone, Peter Albert, Amjad Almahairi, Yasmine Babaei, Nikolay Bashlykov, Soumya Batra, Prajjwal Bhargava, Shruti Bhosale, Dan Bikel, Lukas Blecher, Cristian~Canton Ferrer, Moya Chen, Guillem Cucurull, David Esiobu, Jude Fernandes, Jeremy Fu, Wenyin Fu, Brian Fuller, Cynthia Gao, Vedanuj Goswami, Naman Goyal, Anthony Hartshorn, Saghar Hosseini, Rui Hou, Hakan Inan, Marcin Kardas, Viktor Kerkez, Madian Khabsa, Isabel Kloumann, Artem Korenev, Punit~Singh Koura, Marie-Anne Lachaux, Thibaut Lavril, Jenya Lee, Diana Liskovich, Yinghai Lu, Yuning Mao, Xavier Martinet, Todor Mihaylov, Pushkar Mishra, Igor Molybog, Yixin Nie, Andrew Poulton, Jeremy Reizenstein, Rashi Rungta, Kalyan Saladi, Alan Schelten, Ruan Silva, Eric~Michael Smith, Ranjan Subramanian, Xiaoqing~Ellen Tan, Binh Tang, Ross Taylor, Adina Williams, Jian~Xiang Kuan, Puxin Xu, Zheng Yan, Iliyan Zarov, Yuchen Zhang, Angela Fan, Melanie Kambadur, Sharan Narang, Aurelien Rodriguez, Robert Stojnic, Sergey Edunov, and Thomas
  Scialom. 2023.
\newblock \href {https://arxiv.org/abs/2307.09288} {Llama 2: Open foundation and fine-tuned chat models}.
\newblock \emph{Preprint}, arXiv:2307.09288.

\bibitem[{Wang et~al.(2024{\natexlab{a}})Wang, Zhang, Du, Zhang, and Chu}]{wang2024survey}
Jiahao Wang, Bolin Zhang, Qianlong Du, Jiajun Zhang, and Dianhui Chu. 2024{\natexlab{a}}.
\newblock A survey on data selection for llm instruction tuning.
\newblock \emph{arXiv preprint arXiv:2402.05123}.

\bibitem[{Wang et~al.(2024{\natexlab{b}})Wang, Jain, Zhang, Ray, Kumar, and Athiwaratkun}]{wang2024reasoning}
Junlin Wang, Siddhartha Jain, Dejiao Zhang, Baishakhi Ray, Varun Kumar, and Ben Athiwaratkun. 2024{\natexlab{b}}.
\newblock Reasoning in token economies: Budget-aware evaluation of llm reasoning strategies.
\newblock \emph{arXiv preprint arXiv:2406.06461}.

\bibitem[{Wang et~al.(2019)Wang, Ma, and Meng}]{8788644}
Ling Wang, Qian Ma, and Jianyao Meng. 2019.
\newblock \href {https://doi.org/10.1109/ACCESS.2019.2933361} {Incremental fuzzy association rule mining for classification and regression}.
\newblock \emph{IEEE Access}, 7:121095--121110.

\bibitem[{Wei et~al.(2022{\natexlab{a}})Wei, Bosma, Zhao, Guu, Yu, Lester, Du, Dai, and Le}]{wei2022finetuned}
Jason Wei, Maarten Bosma, Vincent Zhao, Kelvin Guu, Adams~Wei Yu, Brian Lester, Nan Du, Andrew~M. Dai, and Quoc~V Le. 2022{\natexlab{a}}.
\newblock \href {https://openreview.net/forum?id=gEZrGCozdqR} {Finetuned language models are zero-shot learners}.
\newblock In \emph{International Conference on Learning Representations}.

\bibitem[{Wei et~al.(2022{\natexlab{b}})Wei, Wang, Schuurmans, Bosma, brian ichter, Xia, Chi, Le, and Zhou}]{wei2022chain}
Jason Wei, Xuezhi Wang, Dale Schuurmans, Maarten Bosma, brian ichter, Fei Xia, Ed~H. Chi, Quoc~V Le, and Denny Zhou. 2022{\natexlab{b}}.
\newblock \href {https://openreview.net/forum?id=_VjQlMeSB_J} {Chain of thought prompting elicits reasoning in large language models}.
\newblock In \emph{Advances in Neural Information Processing Systems}.

\bibitem[{Xia et~al.(2024)Xia, Li, Liu, Wu, and Liu}]{xia2024evaluating}
Shijie Xia, Xuefeng Li, Yixin Liu, Tongshuang Wu, and Pengfei Liu. 2024.
\newblock Evaluating mathematical reasoning beyond accuracy.
\newblock \emph{arXiv preprint arXiv:2404.05692}.

\bibitem[{Xu et~al.(2024)Xu, Sun, Zheng, Geng, Zhao, Feng, Tao, Lin, and Jiang}]{xu2024wizardlm}
Can Xu, Qingfeng Sun, Kai Zheng, Xiubo Geng, Pu~Zhao, Jiazhan Feng, Chongyang Tao, Qingwei Lin, and Daxin Jiang. 2024.
\newblock \href {https://openreview.net/forum?id=CfXh93NDgH} {Wizard{LM}: Empowering large pre-trained language models to follow complex instructions}.
\newblock In \emph{The Twelfth International Conference on Learning Representations}.

\bibitem[{Yager and Zadeh(1992)}]{fuzzylogic}
Ronald~R. Yager and Lotfi~A. Zadeh. 1992.
\newblock \emph{An Introduction to Fuzzy Logic Applications in Intelligent Systems}.
\newblock Kluwer Academic Publishers, USA.

\bibitem[{Young et~al.(2024)Young, Chen, Li, Huang, Zhang, Zhang, Li, Zhu, Chen, Chang et~al.}]{young2024yi}
Alex Young, Bei Chen, Chao Li, Chengen Huang, Ge~Zhang, Guanwei Zhang, Heng Li, Jiangcheng Zhu, Jianqun Chen, Jing Chang, et~al. 2024.
\newblock Yi: Open foundation models by 01. ai.
\newblock \emph{arXiv preprint arXiv:2403.04652}.

\bibitem[{Yu et~al.(2023{\natexlab{a}})Yu, Zhang, Tiwari, and Wang}]{yu2023natural}
Fei Yu, Hongbo Zhang, Prayag Tiwari, and Benyou Wang. 2023{\natexlab{a}}.
\newblock Natural language reasoning, a survey.
\newblock \emph{ACM Computing Surveys}.

\bibitem[{Yu et~al.(2023{\natexlab{b}})Yu, Jiang, Shi, Yu, Liu, Zhang, Kwok, Li, Weller, and Liu}]{yu2023metamath}
Longhui Yu, Weisen Jiang, Han Shi, Jincheng Yu, Zhengying Liu, Yu~Zhang, James~T Kwok, Zhenguo Li, Adrian Weller, and Weiyang Liu. 2023{\natexlab{b}}.
\newblock Metamath: Bootstrap your own mathematical questions for large language models.
\newblock \emph{arXiv preprint arXiv:2309.12284}.

\bibitem[{Yue et~al.(2023)Yue, Qu, Zhang, Fu, Huang, Sun, Su, and Chen}]{yue2023mammoth}
Xiang Yue, Xingwei Qu, Ge~Zhang, Yao Fu, Wenhao Huang, Huan Sun, Yu~Su, and Wenhu Chen. 2023.
\newblock Mammoth: Building math generalist models through hybrid instruction tuning.
\newblock \emph{arXiv preprint arXiv:2309.05653}.

\bibitem[{Zadeh(1968)}]{Zadeh1968ProbabilityMO}
Lotfi~A. Zadeh. 1968.
\newblock \href {https://api.semanticscholar.org/CorpusID:120966792} {Probability measures of fuzzy events}.
\newblock \emph{Journal of Mathematical Analysis and Applications}, 23:421--427.

\bibitem[{Zhang et~al.(2024{\natexlab{a}})Zhang, Zhou, Wei, Zhao, Sha, Wang, and Wen}]{zhang2024evaluating}
Beichen Zhang, Kun Zhou, Xilin Wei, Xin Zhao, Jing Sha, Shijin Wang, and Ji-Rong Wen. 2024{\natexlab{a}}.
\newblock Evaluating and improving tool-augmented computation-intensive math reasoning.
\newblock \emph{Advances in Neural Information Processing Systems}, 36.

\bibitem[{Zhang et~al.(2023)Zhang, Dong, Li, Zhang, Sun, Wang, Li, Hu, Zhang, Wu et~al.}]{zhang2023instruction}
Shengyu Zhang, Linfeng Dong, Xiaoya Li, Sen Zhang, Xiaofei Sun, Shuhe Wang, Jiwei Li, Runyi Hu, Tianwei Zhang, Fei Wu, et~al. 2023.
\newblock Instruction tuning for large language models: A survey.
\newblock \emph{arXiv preprint arXiv:2308.10792}.

\bibitem[{Zhang et~al.(2024{\natexlab{b}})Zhang, Chen, Ye, Yang, Chen, Wang, and Petzold}]{zhang2024unveiling}
Xinlu Zhang, Zhiyu~Zoey Chen, Xi~Ye, Xianjun Yang, Lichang Chen, William~Yang Wang, and Linda~Ruth Petzold. 2024{\natexlab{b}}.
\newblock \href {https://arxiv.org/abs/2405.20535} {Unveiling the impact of coding data instruction fine-tuning on large language models reasoning}.
\newblock \emph{Preprint}, arXiv:2405.20535.

\bibitem[{Zheng et~al.(2023)Zheng, Liu, Xie, Li, and Li}]{zheng2023progressive}
Chuanyang Zheng, Zhengying Liu, Enze Xie, Zhenguo Li, and Yu~Li. 2023.
\newblock Progressive-hint prompting improves reasoning in large language models.
\newblock \emph{arXiv preprint arXiv:2304.09797}.

\bibitem[{Zhou et~al.(2023)Zhou, Wang, Lu, Shi, Luo, Qin, Lu, Jia, Song, Zhan et~al.}]{zhou2023solving}
Aojun Zhou, Ke~Wang, Zimu Lu, Weikang Shi, Sichun Luo, Zipeng Qin, Shaoqing Lu, Anya Jia, Linqi Song, Mingjie Zhan, et~al. 2023.
\newblock Solving challenging math word problems using gpt-4 code interpreter with code-based self-verification.
\newblock \emph{arXiv preprint arXiv:2308.07921}.

\bibitem[{Zhou and Zhao(2024)}]{zhou2024dual}
Yongwei Zhou and Tiejun Zhao. 2024.
\newblock Dual instruction tuning with large language models for mathematical reasoning.
\newblock \emph{arXiv preprint arXiv:2403.18295}.

\bibitem[{Zhu et~al.(2024)Zhu, Chen, Wang, Gong, Yang, and Xie}]{zhu2024dyval}
Kaijie Zhu, Jiaao Chen, Jindong Wang, Neil~Zhenqiang Gong, Diyi Yang, and Xing Xie. 2024.
\newblock \href {https://openreview.net/forum?id=gjfOL9z5Xr} {Dyval: Dynamic evaluation of large language models for reasoning tasks}.
\newblock In \emph{The Twelfth International Conference on Learning Representations}.

\end{thebibliography}

\appendix

\section{\ourbenchmark Performance}
\label{sec: FRoG performance}
The performance of \ourbenchmark on three masking strategies is listed in Table~\ref{tab: FRoG performance}, the best performing model, GPT-4-turbo has below 50\% accuracy across different settings. The results of inverse scaling effect are summarized in Table~\ref{tab: inverse scaling} where one check represents an inverse scaling effect observed.

\begin{table*}[ht]
    \small
    \centering
    \begin{tabular}{p{3cm}|p{1.2cm}|c|c|c}
    \toprule
    \Thead{Model} & \Thead{\#Param.} & \Thead{Mask} & \Thead{Mislead} & \Thead{X\%} \\
    \midrule
    GPT-4-turbo & - & 33.7 / 48.1 & 34.7 / 44.0 & 37.5 / 49.8 \\
    \midrule
    GPT-3.5-turbo-1106 & - & 20.4 / 31.2 & 22.6 / 27.8 & 23.5 / 31.6 \\
    \midrule
    Llama-2-7b & 7B & 22.5 / 21.9 & 21.5 / 22.8 & 22.4 / 23.7 \\
    % \midrule
    Llama-2-7b-Chat & 7B & 24.3 / 26.8 & 26.9 / 24.7 & 23.7 / 25.0 \\
    \midrule
    Llama-2-13b & 13B & 23.8 / 25.4 & 23.9 / 24.7 & 22.7 / 23.2 \\
    % \midrule
    Llama-2-13b-Chat & 13B & 25.6 / 26.6 & 23.9 / 24.8 & 26.0 / 27.7 \\
    \midrule
    Llama-2-70b & 70B & 24.0 / 27.5 & 23.2 / 25.7 & 23.2 / 27.4 \\
    % \midrule
    Llama-2-70b-Chat & 70B & 25.3 / 31.9 & 26.5 / 29.8 & 26.5 / 28.5 \\
    \midrule
    CodeLlama-7b & 7B & 21.7 / 23.0 & 21.7 / 21.7 & 24.4 / 23.8 \\
    % \midrule
    CodeLlama-34b & 34B & 22.6 / 26.9 & 21.6 / 23.9 & 23.6 / 26.3 \\
    % \midrule
    CodeLlama-70b & 70B & 7.5 / 8.1 & 6.6 / 7.2 & 7.0 / 7.8 \\
    \midrule
    Llama-3-8b & 8B & 22.9 / 26.6 & 21.9 / 24.4 & 23.8 / 24.9 \\
    % \midrule
    Llama-3-8b-Instruct & 8B & 20.7 / 22.8 & 16.9 / 17.9 & 18.9 / 23.0 \\
    \midrule
    Llama-3-70b & 70B & 27.3 / 36.0 & 23.2 / 29.9 & 28.0 / 33.8 \\
    % \midrule
    Llama-3-70b-Instruct & 70B & 28.0 / 42.9 & 25.7 / 39.0 & 29.0 / 42.1 \\
    \midrule
    Llemma-7b & 7B & 23.5 / 23.4 & 23.4 / 22.4 & 23.6 / 24.2 \\
    Llemma-34b & 34B & 25.0 / 28.6 & 23.9 / 23.2 & 24.6 / 27.9 \\
    % \midrule
    \midrule
    Mistral-7b & 7B & 24.7 / 25.0 & 22.3 / 25.0 & 24.1 / 25.2 \\
    % \midrule
    Mixtral-8x7b & 56B & 23.4 / 29.1 & 25.1 / 27.2 & 24.8 / 28.8 \\
    \midrule
    Olmo-1b & 1B & 18.5 / 17.9 & 18.1 / 18.0 & 17.6 / 18.3 \\
    % \midrule
    Olmo-7b & 7B & 22.7 / 22.6 & 21.3 / 20.6 & 21.5 / 23.2 \\
    \midrule
    Qwen-1.5-1.8b & 1.8B & 19.1 / 20.9 & 17.6 / 18.8 & 21.0 / 20.2 \\
    % \midrule
    Qwen-1.5-1.8b-Chat & 1.8B & 21.4 / 19.1 & 20.3 / 18.9 & 21.5 / 19.3 \\
     \midrule
    Qwen-1.5-4b-Chat & 4B & 15.5 / 18.3 & 19.2 / 18.3 & 17.7 / 19.1 \\
    \midrule
    Qwen-1.5-7b & 7B & 21.5 / 26.3 & 22.1 / 25.0 & 23.2 / 27.5 \\
    % \midrule
    Qwen-1.5-7b-Chat & 7B & 20.7 / 25.1 & 18.3 / 21.5 & 21.7 / 24.4 \\
    \midrule
    Qwen-1.5-14b & 14B & 23.3 / 28.6 & 23.2 / 29.4 & 24.0 / 29.6 \\
    % \midrule
    Qwen-1.5-14b-Chat & 14B & 23.6 / 32.1 & 23.7 / 29.0 & 25.8 / 32.5 \\
    \midrule
    Qwen-1.5-32b & 32B & 26.5 / 29.7 & 21.8 / 25.3 & 25.4 / 30.4 \\
    % \midrule
    Qwen-1.5-32b-Chat & 32B & 25.4 / 32.6 & 19.1 / 22.9 & 24.5 / 32.4 \\
   \midrule
    Qwen-1.5-72b-Chat & 72B & 26.4 / 31.8 & 22.3 / 25.6 & 25.1 / 34.1 \\
    \midrule
    Tulu-2-7b & 7B & 6.6 / 6.2 & 8.5 / 6.4 & 7.6 / 7.3 \\
    Tulu-2-DPO-7b & 7B & 24.6 / 28.2 & 23.4 / 24.9 & 23.9 / 26.1 \\
    \midrule
    Tulu-2-13b & 13B & 22.2 / 24.2 & 18.8 / 21.6 & 21.4 / 24.2 \\
    Tulu-2-DPO-13b & 13B & 25.6 / 26.2 & 20.3 / 21.3 & 24.1 / 25.2 \\
    \midrule
    Tulu-2-70b & 70B & 25.3 / 26.6 & 23.7 / 23.5 & 26.2 / 28.9 \\
    Tulu-2-DPO-70b & 70B & 26.3 / 29.1 & 21.4 / 26.0 & 24.9 / 29.6 \\
    \midrule
    WizardLM-7b$^{*}$ & 7B & 21.0 / 21.6 & 18.2 / 17.6 & 19.9 / 20.7 \\
    WizardMath-7b$^{*}$ & 7B & 18.0 / 20.1 & 17.6 / 18.2 & 18.2 / 17.4 \\
    \midrule
    WizardMath-13b$^{*}$ & 13B & 21.6 / 22.4 & 20.4 / 21.4 & 21.9 / 23.4 \\
    \midrule
    WizardLM-70b$^{*}$ & 70B & 24.0 / 26.7 & 23.6 / 26.0 & 25.1 / 27.4 \\
    WizardMath-70b$^{*}$ & 70B & 22.8 / 24.3 & 20.1 / 20.8 & 21.0 / 22.5 \\
    \midrule
    Yi-6b-Chat & 6B & 23.4 / 25.4 & 20.4 / 21.3 & 20.9 / 24.5 \\
    Yi-34b-Chat & 34B & 20.0 / 27.5 & 24.0 / 28.3 & 22.2 / 26.9 \\
    % \midrule
    \bottomrule
    \end{tabular}
    \caption{Performance on \ourbenchmarknospace. $^{*}$number of demonstrations is reduced to 3 avoid outputs truncated by length limits.}
    \label{tab: FRoG performance}
\end{table*}

\begin{table*}[ht]
    \small
    \centering
    \begin{tabular}{p{2.2cm}|p{1.5cm}|l|l|l}
    \toprule
    \Thead{Model} & \Thead{Backbone} & \Thead{[MASK]} & \Thead{Mislead} & \Thead{X\%} \\
    \midrule
    Tulu-2 & Llama &  &  &  \\
    Tulu-2-DPO & Llama & \ding{51} \ding{51} & \ding{51} \ding{51} & \ding{51} \ding{51} \\
    WizardLM & Llama &  &  &  \\
    WizardMath & Llama  & \quad \ding{51} & \ding{51} \ding{51} & \ding{51} \ding{51} \\
    Llama-2 & Llama-2 & \ding{51} & \ding{51} & \ding{51} \ding{51} \\
    Llama-2-Chat & Llama-2 & \quad \ding{51} & \ding{51} \ding{51} & \ding{51} \ding{51} \\
    CodeLlama & Llama-2 &  \ding{51} \ding{51} &  \ding{51} \ding{51} &  \ding{51} \ding{51} \\
    Llemma & Llama-2 &  & \ding{51} \ding{51} &  \\
    Llama-3 & Llama-3 &  &  &  \\
    Llama-3-Instruct & Llama-3 &  &  &  \\
    Mistral & Mistral & \ding{51} &  & \ding{51} \\
    Olmo & Olmo &  &  &  \\
    Qwen-1.5 & Qwen-1.5 &  & \ding{51} \ding{51} & \quad \ding{51} \\
    Qwen-1.5-Chat & Qwen-1.5 & \ding{51} \ding{51} & \ding{51} \ding{51} & \ding{51} \ding{51} \\
    Yi-Chat & Yi & \ding{51} &  &  \\
    \midrule
    \end{tabular}
    \caption{Inverse scaling checkbox, the two checks represent inverse scaling effect (less than 1 point performance gain by scaling model sizes) observed on \benchmarkeasy \Hquad (left) and \benchmarkhard \Hquad (right) respectively.}
    \label{tab: inverse scaling}
\end{table*}

\section{Additional Case Studies}
\label{sec: more example in fuzzy reasoning mechansims}
We list additional examples from GPT-4-turbo on \benchmarkhard in Table~\ref{tab: FRoG examples fuzzy mechanisms extra}, where Ex1, Ex2 and Ex3 represents the three primary fuzzy reasoning strategies in Section~\ref{sec: case study} respectively. For example, in Ex2, the model solves the target percentage mention with 30\% instead of 20\%, but still select the correct quantifier. And in Ex3, the model skips the step of dividing 70 by 127 and directly estimate the GQ preference from it.

\begin{table*}[ht]
    \small
    \centering
    \begin{tabular}{p{15cm}}
    \toprule        
        Ex1: From the sale of sleeping bags, a retailer made a gross profit of [MASK] of the wholesale cost. If each sleeping bag was sold for \$28, what was the wholesale cost per bag?

        If the answer to the question is 24.78, then please select the quantifier that is closest to the meaning of [MASK] from the following choices. [13\%] \\
        A. some \underline{B. small amount} C. few D. tiny amount\\
        \midrule
        Suppose the gross profit percentage is x and the wholesale cost is W per bag. Then the selling price (SP) per bag is \$28, and the actual relation can be established using:
        SP = W + x\% of W
        28 = W + (x/100) * W 
        28 = W(1 + x/100)
        W = 28 / (1 + x/100)
        
        Given that W is \$24.78,
        24.78 = 28 / (1 + x/100)
        1 + x/100 = 28 / 24.78
        x/100 = (28 / 24.78) - 1
        x/100 = 1.13 - 1
        x/100 = 0.13
        x = 13
        
        Therefore, the gross profit percentage is 13\%. This percentage is a moderate amount, not minimal, nor very large.
        
        \quantfuzzyreason{Checking the choices, none represent precisely '13\%', but thinking about its relative value:}\\
        \quantfuzzyreason{- `some' generally suggests a moderate, indefinite amount}\\
        \quantfuzzyreason{- `small amount', `few', and `tiny amount' suggest smaller proportions}
        
        Thus, among all the choices, \textbf{'some' is the best fit} for representing a gross profit of 13\% of the wholesale cost. \\
        \midrule
        Ex2: A shopkeeper has 280 kg of apples. He sells 30\% of these at 20\% profit and the remaining 60\% at [MASK] profit. Find his \% profit on total. If the answer to the question is 24\%, then please select the quantifier that is closest to the meaning of [MASK] from the following choices. [20\%] \\
        A. some \underline{B. moderate amount} C. small amount D. few\\
        \midrule
        The shopkeeper has 280 kg of apples. 30\% of these apples sold at 20\% profit means 0.3 * 280 = 84 kg sold at 20\% profit.
        
        The remaining 60\% sold at an unknown percentage, x, of profit is 0.6 * 280 = 168 kg.
        
        To find the total profit percentage on the whole 280 kg:
        
        1. Calculation for profit from the first 84 kg:
           Profit = 20\% of 84 kg = 0.2 * 84 = 16.8 kg equivalent profit.
        
        2. Calculation for profit from the remaining 168 kg:
           Profit = x\% of 168 kg = 0.01x * 168 = 1.68x kg equivalent profit.
        
        Now, total profit from 280 kg = (16.8 + 1.68x) kg.
        
        Total profit percentage = (Total profit / Total kg) * 100
                               = [(16.8 + 1.68x) / 280] * 100
                               = (0.06 + 0.006x) * 100
                               = 6 + 0.6x.
        
        Given the problem, the total profit percentage is 24\%:
        6 + 0.6x = 24
        0.6x = 24 - 6
        0.6x = 18
        x = 18 / 0.6
        x = 30\%.
        
        So, the remaining 60\% of apples were sold at 30\% profit. Looking at the choices given for the interpretation of [MASK], the closest to a 30\% profit level among the options provided is \textbf{"moderate amount"}, considering it reflects a sizable yet not excessively high-profit margin. \\
        \midrule
        Ex3: A candidate appearing for an examination has to secure [MASK] marks to pass paper i. But he secured only 45 marks and failed by 25 marks. What is the maximum mark for paper i? If the answer to the question is 127, then please select the quantifier that is closest to the meaning of [MASK] from the following choices. [55\%]  \\
        A. moderate amount B. some C. all \underline{D. most}\\
        \midrule
        The candidate secured 45 marks and failed by 25 marks. This means the passing mark for paper I is 45 marks + 25 marks = 70 marks.
        
        Since the maximum mark for paper I is given to be 127, the passing mark (70 marks) out of the maximum (127 marks) represents more than 50\% but significantly less than 100\%. This suggests a quantifier that indicates a majority but not the entirety or something overly general.
        
        \quantfuzzyreason{Among the choices:}\\
        \quantfuzzyreason{A. moderate amount - This generally indicates a reasonable or average proportion, not specifically majority.} \\
        \quantfuzzyreason{B. some - Generally refers to an unspecified quantity, not necessarily a majority.}\\
        \quantfuzzyreason{C. all - This would indicate the entirety, which is not suitable here as 70 is less than 127.} \\
        \quantfuzzyreason{D. most - This generally refers to the majority or the largest part but not the entirety, fitting the criterion where 70 is more than half of 127 but not all of it.}\\
        
        Therefore, `most' is the quantifier that is closest to the meaning required for the passing marks in the context given. \textbf{\#\#\#\# D} \\
    \bottomrule
    \end{tabular}
    \caption{Extra sampled results from GPT-4-turbo in \ourbenchmarknospace, where \quantfuzzyreason{explicit quantifier estimation} is highlighted.}
    \label{tab: FRoG examples fuzzy mechanisms extra}
\end{table*}

\section{Quantifier Understanding Across Scale}
\label{sec: quantifier understanding across scale}
We demonstrate examples from Qwen-1.5-Chat in \benchmarkhard where models of different number of parameters hold different understanding of GQ semantics in Table~\ref{tab: FRoG different quant choice}. Take Ex1 for example, The 7B, 14B and 72B version of Qwen-1.5-Chat all compute the target target mention 0.5 correctly, but reaching to different GQ preferences (\textit{small amount, moderate amount and some}) regarding interpreting the target percentage value. We list the exploration of aligning model behavior to specific quantifier interpretation as future work.

\begin{table*}[ht]
    \small
    \centering
    \begin{tabular}{p{15cm}}
    \toprule     
        \textbf{Ex1} Mary can do a piece of work in 12 days. rosy is [MASK] more efficient than mary. the number of days taken by rosy to do the same piece of work is? If the answer to the question is 8, then please select the quantifier that is closest to the meaning of [MASK] from the following choices. [50\%]\\
        \underline{A. moderate amount} B. some C. most D. small amount \\
        \midrule
        \small{\textbf{[Qwen-1.5-72B-Chat]}} If Mary takes 12 days to complete the work, Rosy, who is more efficient, will take less time. Let's say Mary's efficiency is 1 unit of work per day, then Rosy's efficiency is Mary's plus [MASK] units. If Rosy is `x' times more efficient than Mary, her efficiency would be (1+x) units per day. \\

        Since the work is the same for both, we can equate the efficiency in terms of time:\\
        Mary: 1 unit/day * 12 days = 12 units\\
        Rosy: (1+x) units/day * d days = 12 units\\
        
        Solving for d, we have
        12 / (1 + x) = d\\
        
        Given the answer is 8 days for d, we can plug this value back into the equation to find x:\\
        12 / (1 + x) = 8\\
        1 + x = 12 / 8\\
        1 + x = 1.5\\
        x = 0.5\\
        
        Since 0.5 represents a "small amount" more efficiency, the answer is \textbf{D. small amount. \#\#\#\# D}\\
        \midrule
        \small{\textbf{[Qwen-1.5-14B-Chat]}} Suppose Mary's work efficiency is `M' days' work. Then, Rosy's efficiency is `M + [MASK]' days' work. Given, Rosy can complete the work in 'X' days.\\

        So, (M + [MASK]) * X = M * 12\\
        Given X = 8 days, we can find what [MASK] represents:
        
        12M / X = M + [MASK]
        
        12 / 8 = 1 + [MASK]
        
        [MASK] = 12 / 8 - 1
        
        [MASK] = 3/2 - 1
        
        [MASK] = 1/2
        
        This means Rosy is half as efficient as Mary, or in other words, she takes twice the time. \quantfuzzyreason{This corresponds to a "moderate amount" of difference because, compared to `most' or a large amount, which would be a huge difference, a small amount wouldn't result in Rosy taking twice as much time. The other options are `some' or `tiny amount', which are not as accurate as `moderate amount' in this context because they could mean a lesser extent of efficiency difference.} \\
        
        Therefore, the answer is \textbf{A. moderate amount} \\
        \midrule
        \small{\textbf{[Qwen-1.5-7B-Chat]}} Rosy is `x' times more efficient than Mary. So, the time taken by Rosy is 12 / (1 + x) days.\\
        We are given that 12 / (1 + x) = 8.\\
        Let's solve for x: (1 + x) = 12 / 8 = 1.5\\
        Simplifying x, x = 0.5.\\
        This means that Rosy is half as efficient as Mary. So, `some' is the closest quantifier choice for `most'. If x were to be a tiny portion, this would not make sense.
        Therefore, the answer is \textbf{`some' (or B). \#\#\#\# B} \\
    \bottomrule
    \end{tabular}
    \caption{Examples of Qwen-1.5-Chat models on \ourbenchmark where models of different parameters compute similar percentage values about the target percentage mention, but holding different opinions about the quantifier choice. The \quantfuzzyreason{explicit quantifier estimation} is highlighted.}
    \label{tab: FRoG different quant choice}
\end{table*}

\section{\ourbenchmark Templates}
\label{sec: FRoG templates}
We list the \ourbenchmark templates employed in Table~\ref{tab: FRoG template}.

\begin{table*}[ht!]
\begin{tcolorbox}[colback=blue!5!white,colframe=black,width=1.0\textwidth,title={Prompt Templates for \ourbenchmark Problems}]
\small
Given the (Question, Original Answer) of the original benchmark, the \ourbenchmark problems can be constructed with the following templates.\\
\newline
\#\#\# Problem Template\\
Question: \\
\{FRoG Question\}\\
Answer:\\
Let's think step by step\\

\#\#\# FRoG Question Template\\
\text{[}MASK\text{]}: \\
\{Question\}\\
\{Choices\}\\
If the answer to the question is \{Original Answer\}, then please select the quantifier that is closest to the meaning of [MASK] from the following choices.\\
\newline
\text{[}Mislead\text{]}: \\
\{Question\}\\
\{Choices\}\\
If the answer to the question is \{Original Answer\}, and \{Mislead Quantifier\} in the question is not correct, please select the most appropriate quantifier from the following choices.\\
\newline
\text{[}X\%\text{]}: \\
\{Question\}\\
\{Choices\}\\
If the answer to the question is \{Original Answer\}, then please select the most appropriate quantifier that is closest to the meaning of X\% from the following choices.\\
\newline
\text{[}Mask Percentage\text{]}: \\
\{Question\}\\
\{Choices\}\\
If the answer to the question is \{Original Answer\}, then please select the percentage value that is closest to the meaning of [MASK] from the following choices.\\
\end{tcolorbox}
\caption{Prompt template for \ourbenchmark problems.}
\label{tab: FRoG template}
\end{table*}

\section{Instruction}
\label{sec: FRoG instruction}
We list the \ourbenchmark instruction employed in Table~\ref{tab: FRoG instruction}.

\begin{table*}[ht!]
\begin{tcolorbox}[colback=blue!5!white,colframe=black,width=1.0\textwidth,title={Instruction used for \ourbenchmark Evaluation}]
\small
You are an expert in mathematical reasoning and generalized quantifier reasoning. Here you are asked to answer one mathematical question based on real-life scenarios with a description starting with 'Question:' For example, the question may describe the driving experience of a person. Your answer will start with 'Answer: let's think step by step'. \\

You will also be provided with four possible choices, please select the choice that is closest to your estimation of the answer.\\

The answer needs to include necessary reasoning steps to demonstrate your thinking procedure, and the final result of your calculation is demonstrated at the end of your answer starting with `\#\#\#\#'.\\ 

Here are some examples starting with `Question: ' for your reference.
\end{tcolorbox}
\caption{Instruction employed in \ourbenchmark evaluation.}
\label{tab: FRoG instruction}
\end{table*}

\end{document}